\documentclass[10pt,twocolumn,letterpaper]{article}

\usepackage[pagenumbers]{cvpr}

\usepackage{graphicx}
\usepackage{amsmath}
\usepackage{amssymb}
\usepackage{booktabs}
\usepackage{multirow}

\definecolor{cvprblue}{rgb}{0.21,0.49,0.74}
\usepackage[
    breaklinks,
    colorlinks,
    allcolors=cvprblue
]{hyperref}

\def\confName{CVPR}
\def\confYear{2027}

\title{IGRD: Intent-Guided Representation Disentanglement
for Human-to-Robot Manipulation Transfer}

\author{
    Zhiqian Xia\textsuperscript{1,2}%
    \thanks{Equal contribution. Zhiqian Xia performed this work during an internship at Astribot.}
    \enspace
    Lianyang Ma\textsuperscript{2}\footnotemark[1]
    \enspace
    Siyu Xia\textsuperscript{1}
    \enspace
    Jiancheng Wang\textsuperscript{4}
    \enspace
    Haifeng Xia\textsuperscript{1,3}%
    \thanks{Corresponding author.}
    \\[0.6em]
    \textsuperscript{1}Southeast University
    \quad
    \textsuperscript{2}Astribot
    \\
    \textsuperscript{3}Sun Yat-sen University, Shenzhen, China
    \\
    \textsuperscript{4}Anhui University
}

\begin{document}

\maketitle

\begin{abstract}

Vision-Language-Action (VLA) models have achieved promising performance for generalized robotic manipulation, yet their reliance on expensive real-robot data remains a fundamental bottleneck. A scalable alternative is to transfer knowledge from off-the-shelf human-centric videos. However, existing methods extract latent action representations that are entangled with scene-specific noise, diluting the transferred knowledge. Motivated by this limitation, we propose Intent-Guided Representation Disentanglement (IGRD), a two-stage method that transfers knowledge through a clean action representation shared across humans and robots. In the first stage, IGRD extracts explicit action intents and disentangles the implicit action intent from manipulation videos. In the second stage, IGRD aligns the human and robot domains within this shared intent space and decodes the intent into executable robot actions. Extensive experiments in both simulation and real-world environments demonstrate that IGRD transfers human knowledge more effectively and consistently outperforms prior baselines.

\end{abstract}

\section{Introduction}

The rapid progress of large-scale Vision-Language Models (VLMs)~\cite{bai2025qwen2} has established the Vision-Language-Action (VLA) paradigm as a promising route toward generalized robot control~\cite{brohan2022rt}. 
% However, training VLA models typically relies on robotic demonstrations collected through teleoperation. Such a pipeline is labor-intensive, time-consuming, and dependent on expert operators, which inevitably yields datasets of limited diversity and scale. 
However, training VLA models typically relies on robotic demonstrations collected through teleoperation. This data collection process is labor-intensive, time-consuming, and dependent on expert operators, which inevitably yields datasets of limited diversity and scale.

% To alleviate this data bottleneck, a growing body of research has turned to unlabeled human videos. Such videos are ubiquitous and abundant across the internet, and their diverse, real-world interaction scenarios make them a natural source of supervision for robotic learning. Existing human-to-robot learning approaches broadly fall into two categories: Motion Capture (MoCap) based~\cite{kareer2025egomimic,wang2023mimicplay,li2025scalable} and latent action learning~\cite{chen2024igor,bjorck2025gr00t}. MoCap-based typically rely on MoCap-derived hand models to train robot manipulation policies. However, they still depend on specialized capture equipment, so data acquisition remains costly, and the domain gap further limits their generalization. To exploit human videos more directly, latent action learning instead extracts transferable representations from raw videos to guide robot manipulation. Yet these representations are often contaminated by scene noise, which hampers effective knowledge transfer from human videos. Fig.~\ref{figure:first} contrasts these two paradigms with ours along two practical axes: the cost of data acquisition and the cleanliness of the transferred representation.

\begin{figure}[t]
  \centering
  \includegraphics[width=\linewidth]{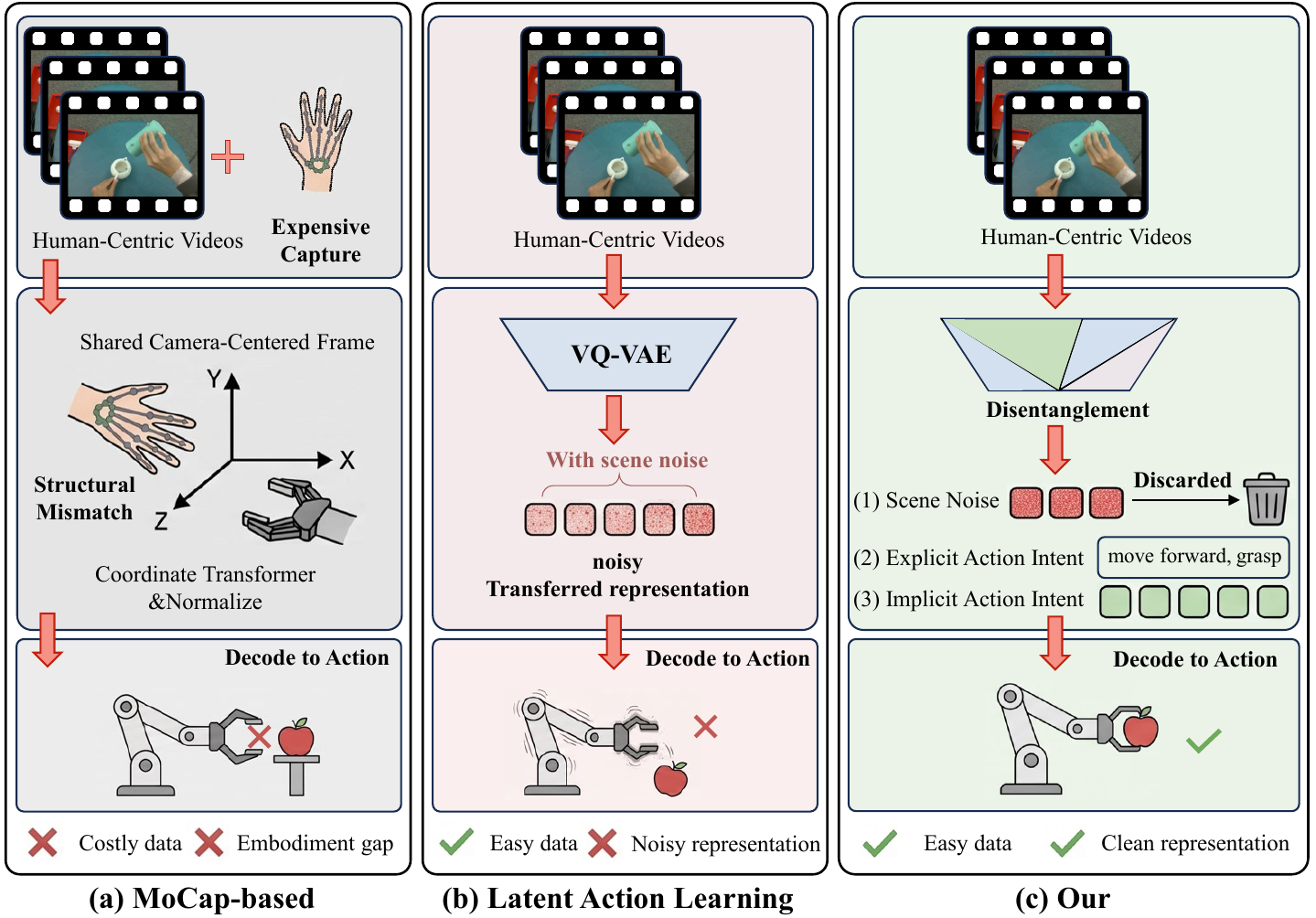}
  \caption{Evolution of human-to-robot learning paradigms. (a) MoCap-based methods depend on specialized capture devices, so data remains costly to collect, and the large embodiment gap limits generalization. (b) Latent action learning is trained directly from abundant raw human videos, but the learned representations are contaminated by scene noise. (c) Our method also learns from easy-to-acquire raw videos, yet extracts a clean Action Intent shared across humans and robots, enabling noise-free knowledge transfer to the robot.}
  \vspace{-0.6em}
  \label{figure:first}
\end{figure}

To alleviate this data bottleneck, a growing body of research has turned to human-centric videos~\cite{luo2026joint,chen2025moto}. Such videos are ubiquitous and abundant across the internet, and their diverse, real-world interaction scenarios make them a natural source of supervision for robotic learning. Existing human-to-robot learning approaches broadly fall into two categories: Motion Capture (MoCap) based~\cite{kareer2025egomimic,wang2023mimicplay,li2025scalable} and latent action learning~\cite{univla,chen2024igor,bjorck2025gr00t}. As shown in Fig.~\ref{figure:first}(a), MoCap-based methods rely on MoCap-derived hand models to train robot manipulation policies. However, they still depend on specialized capture equipment, so data acquisition remains costly, and the embodiment gap between human and robot further limits their generalization. To exploit human videos more directly, latent action learning instead extracts transferable representations from raw videos to guide robot manipulation (Fig.~\ref{figure:first}(b)). Yet these representations are often contaminated by scene noise, which hampers effective knowledge transfer.

Thus, the primary challenge of human-to-robot learning is to \textit{extract clean action representations} from easy-to-acquire human-centric videos and \textit{transfer them to robot manipulation policies}. 
% \textcolor{red}{Inspired by a image tokenization method\cite{languagetoken}, which prove languege discription of a image are a high effecicy compact of the image and thus a image can be tokenizer a language discreption and a discreption irrelevant tokens. Thus as for a video, 也就是从present image to the next image的中间变化状态,can also represent by a language discription and its irrelevant. Thus we could find a sutiable language discription as a bridge across human and robot videos and use it as the signal to guide the representation disentaglement to abtain the clean action rpresentations.}
% Our key observation is that, before acting, humans first break a complex high-level task into an ordered sequence of simple atomic steps. For example, the instruction ``pick up an apple'' naturally decomposes into steps such as ``move forward, grasp''. We formalize this ordered sequence of atomic steps as the \textbf{Action Intent}. Building on this idea, we regard the Action Intent as a reliable bridge between humans and robots.
Inspired by an image tokenization method~\cite{languagetoken},
which shows that a language description provides a compact
summary of an image, we can view an image as a combination of
its language description and residual visual information.
We extend this idea to the transition between two video frames:
a language description can summarize the underlying action,
while residual information captures visual changes not
explained by that description.
Thus, identifying a language description shared by human
and robot videos would provide a common semantic anchor
for separating action-relevant information from scene-
and embodiment-specific variations, facilitating the
learning of clean action representations that transfer
from humans to robots.
To identify such a shared description, we observe that
human and robot manipulations can both be described as
sequences of basic movements and gripper or hand operations,
such as ``move forward'' followed by ``grasp.''
We refer to this description of the intended action as
\emph{action intent}.
By describing what movement to perform without specifying
the appearance of the agent or its surroundings,
action intent provides a shared semantic anchor across
human and robot videos.
As illustrated in Fig.~\ref{figure:first}(c), our method
learns from readily available human-centric videos and
transfers the resulting clean action representations
to robot policies.

To this end, we propose a two-stage framework built around the Action Intent: the Intent-Guided Latent Disentanglement stage extracts a clean Action Intent, and the Action Intent Policy Alignment stage employs it as a bridge to transfer knowledge and ground it into robot actions. 
% In the first stage, we extract explicit action intents (e.g., ``move forward, grasp'') from both human and robot videos, and integrate them with the video inputs to further disentangle implicit action intent tokens, filtering out scene noise to keep the representation clean. To further reduce information overlap between the disentangled representations, we introduce a cross-branch decorrelation regularizer that encourages the implicit action intent to be uncorrelated with the scene representation. 
In the first stage, we extract explicit action intents (e.g., ``move forward, grasp'') from both human and robot videos and use them to guide the separation of implicit action intent tokens from scene noise. To reinforce this separation, we also introduce a cross-branch decorrelation regularizer that reduces redundancy between the implicit action intent and scene representations.
In the second stage, a VLM backbone is jointly pre-trained on human and robot videos within this shared intent space, aligning the two domains, and an action decoder then maps the implicit intent tokens to precise robot actions. Through this design, the Action Intent bridges human videos and robotic VLA models, enabling manipulation knowledge to be transferred as a clean representation shared between humans and robots. The main contributions of this paper are summarized as follows:
\begin{itemize}
    \item We design a method to extract a clean action representation, termed Action Intent, from human-centric videos in both explicit and implicit forms, and transfer it to robot manipulation policies.
    \item We propose a joint pre-training strategy that aligns human and robot data within a shared intent space, enabling effective human-to-robot policy learning.
    \item Extensive experiments in both simulation and real-world settings show that our method consistently outperforms state-of-the-art baselines.
\end{itemize}

\section{Related Work}

\subsection{Vision-language-action Models}

Building upon the robust multimodal reasoning capabilities of Vision-Language Models (VLMs)~\cite{alayrac2022flamingo,liu2023visual}, Vision-Language-Action (VLA) models have emerged to generate robot actions conditioned on visual observations and language instructions. Pioneered by OpenVLA~\cite{kim2024openvla}, some works discretize the action space, enabling VLMs to directly predict action tokens via end-to-end fine-tuning~\cite{zitkovich2023rt}. Conversely, other works explore more robust generation strategies by leveraging VLMs to produce intermediate representations, which are subsequently decoded into robot actions using diffusion models~\cite{li2024cogact,wen2025diffusionvla,liu2025hybridvla} or flow matching~\cite{black2024pi_0,intelligence2504pi0}. More recently, several works~\cite{zawalski2024robotic,zhang2025inspire,zhao2025cot} utilize Chain-of-Thought (CoT) to enhance planning and reasoning capabilities for long-horizon tasks. Nevertheless, these methods remain inherently data-hungry. In contrast, our approach transfers knowledge from human-centric videos to VLAs, effectively alleviating the data scarcity bottleneck.

\subsection{Learning from human videos}

To alleviate data scarcity in VLAs, several studies leverage readily available human videos for robot policy learning~\cite{luo2025being,bu2025agibot,dai2026conla,zhang2026clap,yang2026learning}. MoCap-based methods~\cite{yang2025egovla,qiu2025humanoid} extract explicit human actions but face embodiment gaps. Affordance-based approaches~\cite{bahl2023affordances,chen2025vidbot} capture object-interaction knowledge but lack an explicit bridge for human-to-robot transfer. More recently, methods~\cite{zheng2025flare,lapa} leverage latent actions extracted from videos to pre-train action models. However, task-irrelevant noise and insufficient information extraction can limit their effectiveness. Although UniVLA~\cite{univla} seeks to disentangle task-centric and task-irrelevant representations, information redundancy between the two branches may persist. In contrast, our approach combines intent-guided disentanglement with a cross-branch decorrelation regularizer to reduce such redundancy, yielding cleaner and embodiment-agnostic representations for human-to-robot transfer.

\section{Proposed Method}

\subsection{Preliminaries}
%1把最后一段的中间态重新定义符号，并在method overview里面说明这个符号具体来说包含显式的laten和隐式的向量 all 3

Vision-Language-Action models (VLAs) are designed to synthesize precise robotic actions $a$ conditioned on visual observations $o$ and language instructions $l$. Standard methods follow a direct mapping paradigm $\pi_\theta: (o, l) \rightarrow a$, typically achieved by fine-tuning pre-trained Vision-Language Models. However, this supervised approach is inherently data-hungry and constrained by the scarcity of ground-truth robotic annotations.

To address these data limitations, this paper introduces an Intent-Guided Representation Disentanglement transfer learning strategy leveraging human-centric videos. Concretely, we assume access to a limited dataset of expert robot trajectories, denoted as $\mathcal{D}_{robot} = \{ \tau_i \}_{i=1}^{N}$. Each trajectory $\tau = \{(o_t, l, a_t)\}_{t=1}^{T}$ contains paired visual observations $o_t$, language instructions $l$, and ground-truth actions $a_t$. This dataset supports supervised learning but is expensive to scale. We also utilize a dataset of human-centric videos, denoted as $\mathcal{D}_{human} = \{ v_j \}_{j=1}^{M}$. Unlike robotic data, these videos consist solely of visual frames $v = \{o_t\}_{t=1}^{T}$ without associated action labels or language instructions. The challenge lies in extracting transferable knowledge from $\mathcal{D}_{human}$ to improve the policy $\pi_\theta$ on the target robot domain.

\begin{figure*}[ht]
  \centering
  \includegraphics[width=\linewidth]{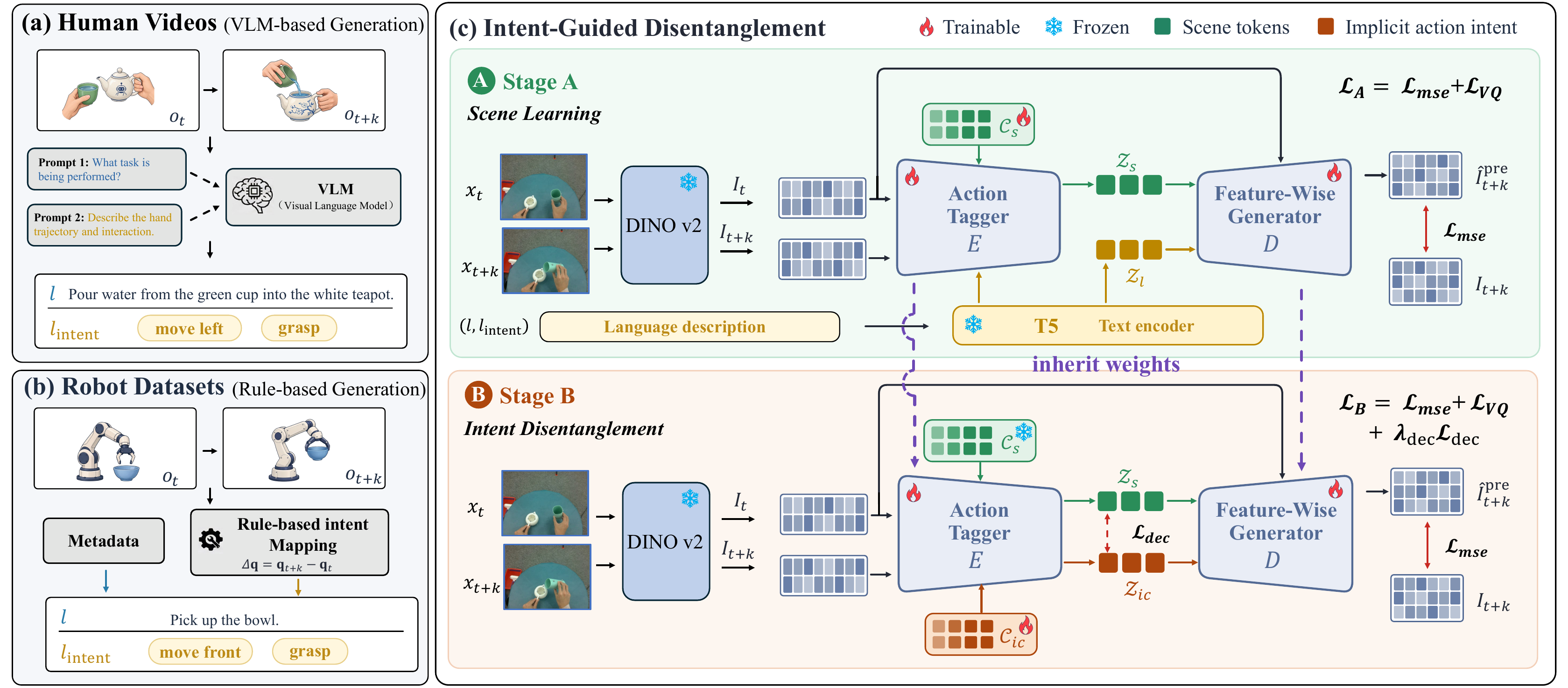}
  % \caption{\textbf{Action-Intent-Guided Latent Disentanglement stage.} This stage comprises: 1) VLM-based and 2) Rule-based Generation, which synthesize high-fidelity language instructions $l$ and explicit action intents $l_{intent}$ for unlabeled human videos and robot datasets, respectively; and 3) Intent-Guided Disentanglement, which extracts the implicit action intent $\mathcal{Z}_{ic}$ from these semantically enriched inputs.}
  % \caption{\textbf{Action-Intent-Guided Latent Disentanglement stage.} It comprises: 1) VLM-based and 2) Rule-based Generation, which synthesize instructions $l$ and explicit intents $l_{intent}$ for human videos and robot data, respectively; and 3) Intent-Guided Disentanglement, which extracts the implicit intent $\mathcal{Z}_{ic}$ from these manipulation videos under intent guidance.}
  % \label{figure:laten}
  \caption{\textbf{Action-Intent-Guided Latent Disentanglement stage.} It comprises: (a) VLM-based Generation, which generates instructions $l$ and explicit intents $l_{\mathrm{intent}}$ for human videos; (b) Rule-based Generation, which derives explicit intents for robot data, with instructions obtained from metadata; and (c) Intent-Guided Disentanglement, which first learns scene tokens under language guidance and then extracts the implicit intent $\mathcal{Z}_{ic}$ with the scene codebook frozen and a cross-branch decorrelation constraint.}
  \vspace{-1.6em}
\label{figure:laten}
\end{figure*}

\subsection{Method Overview}

\begin{figure*}[ht]
  \centering
  \includegraphics[width=\linewidth]{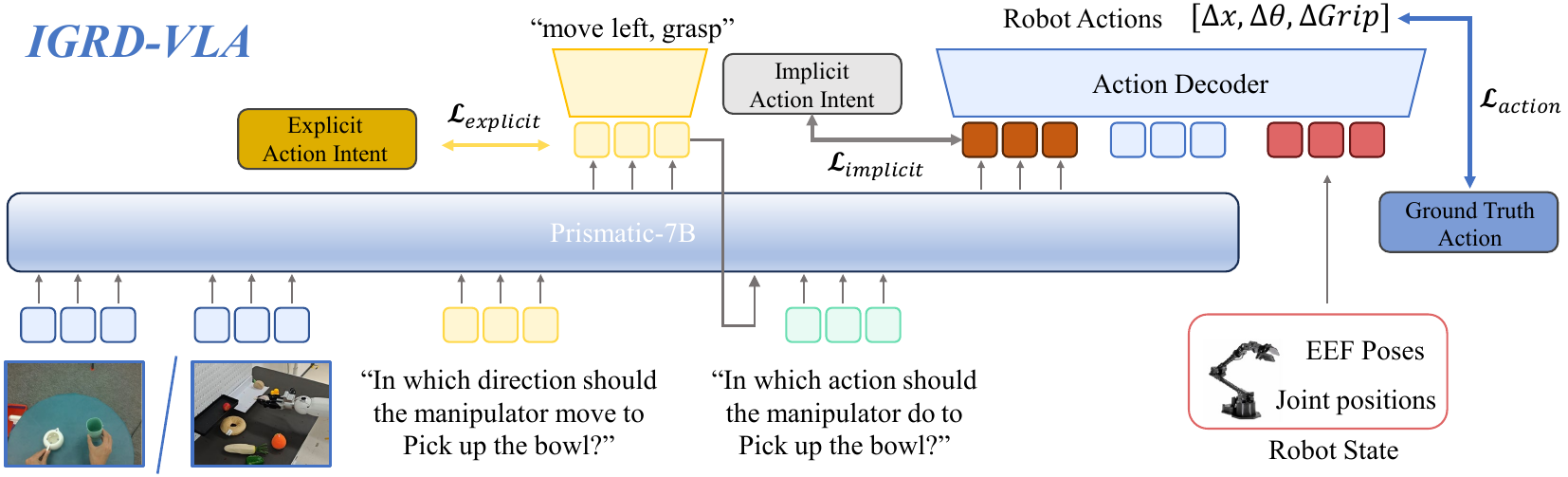}
  \caption{\textbf{Action Intent Policy Alignment stage.} The explicit ($l_{intent}$, yellow) and implicit ($\mathcal{Z}_{ic}$, brown) intents form a shared space that bridges the embodiment gap, allowing our Prismatic-7B backbone to transfer manipulation knowledge from human videos to robots. At deployment, an Action Decoder maps these shared tokens to low-level robot actions.}
  \label{figure:post}
  \vspace{-1em}
\end{figure*}

Human and robot manipulation videos differ in appearance and embodiment, yet often share the same underlying action intents. These shared intents can serve as semantic priors for disentangling action-relevant information from scene variation, promoting representations that transfer more effectively from human videos to robot policies. Concretely, our framework consists of two primary stages, as illustrated in Fig.~\ref{figure:laten} and Fig.~\ref{figure:post}. The Action-Intent-Guided Latent Disentanglement stage aims to obtain a clean action intent that serves as the bridge between the two domains, while the Action Intent Policy Alignment stage leverages this bridge to transfer human knowledge and ground it into executable robot actions.

In the first stage, we extract the action intent $\psi = (l_{intent}, \mathcal{Z}_{ic})$ from both the human dataset $\mathcal{D}_{human}$ and the robot dataset $\mathcal{D}_{robot}$. We first synthesize the explicit intent $l_{intent}$ using a VLM-based strategy for human videos and a rule-based strategy for robot trajectories. By integrating the videos with these explicit intents, we then disentangle the clean implicit intent $\mathcal{Z}_{ic}$ from scene noise, yielding a representation tailored for VLA comprehension:
\begin{equation}
\mathcal{Z}_{ic} = f(o_t, o_{t+k}, l, l_{intent}).
\end{equation}
In the second stage, we align the human and robot representation spaces over the shared intents $(l_{intent}, \mathcal{Z}_{ic})$ through joint pre-training, so that knowledge from human videos flows into the robot policy. An action decoder then maps $\mathcal{Z}_{ic}$ to executable robot actions $a_t$.

\subsection{Action-Intent-Guided Latent Disentanglement}

In this stage, we aim to acquire the action intent $\psi = (l_{intent}, \mathcal{Z}_{ic})$ shared across human and robot videos. Our design follows a coarse-to-fine rationale: \textit{the explicit intent $l_{intent}$ serves as a high-level semantic anchor that guides the disentanglement of the fine-grained implicit intent $\mathcal{Z}_{ic}$}. To obtain $l_{intent}$, we tailor two generation protocols to the differing conditions of the two domains. Human videos lack both language instructions and action labels, so we leverage a VLM to jointly generate the instruction $l$ and the explicit intent $l_{intent}$. While robot trajectories already provide language instructions in their metadata, they still lack intent labels, so we derive $l_{intent}$ directly from the precise proprioceptive states via a rule-based strategy. Guided by these explicit intents, we then introduce an Intent-Guided Disentanglement architecture, comprising an action tagger and a feature-wise generator, to distill the clean implicit intent $\mathcal{Z}_{ic}$ from the videos.

{\bfseries VLM-based Generation (Human Videos).}
This module produces two complementary annotations for each human video. The instruction $l$ is specified as an attribute-rich description that encodes object properties (e.g., color, texture) and spatial relations (e.g., ``Pour the water from the green cup into the left white teapot''), rather than a coarse description (e.g., ``pour water'') that leaves the interacting objects underconstrained; such fine-grained semantics furnish strong supervision for the subsequent disentanglement. The explicit intent $l_{intent}$ is specified as a sequence of discrete motion primitives drawn from a compact vocabulary $\mathcal{A}_{intent}=\{\text{Move Right/Left/Front/Back/Up/Down, Grasp/Release}\}$, forming a high-level, embodiment-agnostic abstraction of the manipulation. We synthesize both annotations by prompting Gemini 2.5 pro~\cite{comanici2025gemini} over each video, deriving $l$ from the visual scene and inferring $l_{intent}$ from the temporal dynamics across frames (prompts detailed in Appendix~\ref{appendix:instruc_prompts}).

Although $\mathcal{A}_{intent}$ is deliberately coarse, the resulting sequence retains sufficient expressiveness to characterize arbitrary manipulations, for two reasons. First, a complex trajectory is temporally partitioned into short steps, each dominated by a single primitive, such that their concatenation recovers the global motion, analogous to approximating a smooth curve with piecewise-linear segments. Second, the fine-grained kinematics within each step (e.g., simultaneous multi-axis motion) are not discarded but preserved in the fine-grained implicit intent $\mathcal{Z}_{ic}$. To ensure label fidelity, all VLM-generated annotations are further verified and corrected by professional annotators, eliminating hallucinations and erroneous primitives; a quantitative reliability audit is reported in Appendix~\ref{appendix:reliability}.

{\bfseries Rule-based Generation (Robot Trajectories).}
Robot trajectories already carry instructions in their metadata, so we compute $l_{intent}$ directly from proprioception. Given the end-effector coordinates at steps $t$ and $t+k$, we form the displacement $\Delta q = [x_{t+k}-x_t,\, y_{t+k}-y_t,\, z_{t+k}-z_t]$ and assign the primitive along its dominant axis (e.g., a dominant $y$ component yields ``move front'' or ``move back'' according to its sign). In parallel, significant changes in gripper aperture are encoded as ``grasp'' or ``release''.

{\bfseries Intent-Guided Disentanglement.}
Given the explicit intents, we extract the implicit intent $\mathcal{Z}_{ic}$ through a two-stage scheme that uses $l_{intent}$ as a conditioning signal to disentangle the intent-relevant motion from the intent-irrelevant scene variation. Concretely, the architecture pairs an Action Tagger $\mathcal{E}$ (a causal spatial-temporal transformer) with a feature-wise generator $\mathcal{D}$ (a spatial transformer), operating in the DINOv2~\cite{dinov2} feature space on $(I_t, I_{t+k})$ with two codebooks $\mathcal{C}_s, \mathcal{C}_{ic} \in \mathbb{R}^{C \times N \times d}$ for the scene variation and the implicit intent, respectively. The two stages progressively strip away the scene noise and then isolate the implicit action intent, as detailed below.

\textbf{Stage A: Scene Learning.} We first let $\mathcal{C}_s$ absorb everything unrelated to the action. Conditioned on the text embedding $\mathcal{Z}_{l}=\text{T5}(l, l_{intent})$~\cite{t5exploring}, which already carries both the task and the coarse intent, the Action Tagger encodes $(I_t, I_{t+k})$ into scene tokens $\mathcal{Z}_{s}$ via $\mathcal{C}_s$, and the generator reconstructs $I_{t+k}$ from $I_t, \mathcal{Z}_{s}, \mathcal{Z}_{l}$. Since the conditioning already explains the intent-relevant motion, $\mathcal{Z}_{s}$ is driven to capture only the intent-irrelevant scene variation (e.g., background, lighting) needed to complete the prediction:
\begin{equation}
\mathcal{Z}_{s} = \mathcal{E}(I_t, I_{t+k}, \mathcal{C}_s, \mathcal{Z}_l), \quad I_{t+k}^{pre} = \mathcal{D}(I_t, \mathcal{Z}_{s}, \mathcal{Z}_{l}).
\end{equation}
% This stage is trained in a self-supervised manner, minimizing the future-frame prediction error together with the vector-quantization loss $\mathcal{L}_{VQ}$~\cite{vqvae_train}:
% \begin{equation}
% \label{eq:vqloss}
% \mathcal{L}_A = ||I_{t+k}^{pre} - I_{t+k}||^2_2 + \mathcal{L}_{VQ}.
% \end{equation}
This stage is trained in a self-supervised manner, minimizing the future-frame prediction error $\mathcal{L}_{\mathrm{mse}} = \|I_{t+k}^{pre} - I_{t+k}\|_2^2$ together with the vector-quantization loss $\mathcal{L}_{VQ}$~\cite{vqvae_train}:
\begin{equation}
\label{eq:vqloss}
\mathcal{L}_A = \mathcal{L}_{\mathrm{mse}} + \mathcal{L}_{VQ}.
\end{equation}

\textbf{Stage B: Intent Disentanglement.}
We initialize this stage with the Stage-A weights, freeze
$\mathcal{C}_s$, and replace the text conditioning with a new
codebook $\mathcal{C}_{ic}$. The generator predicts $I_{t+k}$
from $I_t$, $\mathcal{Z}_s$, and $\mathcal{Z}_{ic}$:
\begin{equation}
\begin{aligned}
\mathcal{Z}_{ic} = \mathcal{E}(I_t, I_{t+k}, \mathcal{C}_s, \mathcal{C}_{ic}), \quad I_{t+k}^{pre} = \mathcal{D}(I_t, \mathcal{Z}_s, \mathcal{Z}_{ic}).
\end{aligned}
\end{equation}
The intent-guided initialization encourages $\mathcal{Z}_s$
to capture scene variation and $\mathcal{Z}_{ic}$ to encode
the complementary intent-relevant motion.

To further reduce redundancy between the two representations,
we introduce a cross-branch decorrelation regularizer.
We flatten each token sequence and standardize each feature
dimension across the mini-batch, obtaining
$\widetilde{S}\in\mathbb{R}^{B\times D_s}$ and
$\widetilde{A}\in\mathbb{R}^{B\times D_a}$ for the scene and
implicit intent representations, respectively.
Here, $B$ is the batch size, and $D_s$ and $D_a$ are the
flattened feature dimensions.
The regularizer is
\begin{equation}
\mathcal{L}_{\mathrm{dec}}
=
\frac{1}{D_sD_a}
\left\|
\frac{1}{B}
\operatorname{sg}(\widetilde{S})^\top \widetilde{A}
\right\|_F^2,
\label{eq:decorrelation}
\end{equation}
where $\operatorname{sg}$ denotes stop-gradient.
Stage B augments the objective in Eq.~\ref{eq:vqloss}
with $\lambda_{\mathrm{dec}}\mathcal{L}_{\mathrm{dec}}$,
encouraging the implicit intent representation to reduce
linear redundancy with the scene representation without
adding inference cost.

\subsection{Action Intent Policy Alignment}

To transfer manipulation knowledge from human videos and adapt it to a specific robotic embodiment, we align the human and robot representation spaces over the shared explicit intent $l_{intent}$ and implicit intent $\mathcal{Z}_{ic}$, and finally map this knowledge to executable robot actions. As illustrated in Fig.~\ref{figure:post}, this alignment proceeds in two phases: a joint pre-training phase that transfers human knowledge into a shared intent space, and a post-training phase that grounds the learned policy on a target robot.

\textbf{Joint Pre-training.} We first learn a unified policy in the shared intent space from both human videos and robot demonstrations, employing Prismatic-7B~\cite{karamcheti2024prismatic} as the backbone $\pi_{\theta}$. Given an observation and instruction, the policy predicts the explicit intent $l_{intent}$ as an intermediate reasoning step and then the implicit intent $\mathcal{Z}_{ic}$:
\begin{equation}
\hat{l}_{intent}, \hat{\mathcal{Z}}_{ic} \sim \pi_{\theta}(l_{intent}, \mathcal{Z}_{ic} \mid \mathbf{O}_t, l).
\end{equation}
To supervise $\mathcal{Z}_{ic}$ autoregressively, we extend the vocabulary with special tokens $\{\texttt{LAT}_1, \dots, \texttt{LAT}_C\}$, where $C$ corresponds to the size of the codebook $\mathcal{C}_{ic}$. The pre-training objective sums the explicit intent loss and the implicit intent loss:
\begin{equation}
\mathcal{L}_{pretrain} = \mathcal{L}_{explicit} + \mathcal{L}_{implicit},
\end{equation}
each being the negative log-likelihood over the explicit intent tokens ($w_i$) and the implicit intent tokens ($z_j$), respectively:
\begin{align}
\mathcal{L}_{\text{explicit}} &= - \sum_{i}^{N_1} \log \pi_{\theta}(w_i \mid \mathbf{O}_t, l, w_{k,<i}) \\
\mathcal{L}_{\text{implicit}} &= - \sum_{j}^{N_2} \log \pi_{\theta}(z_j \mid \mathbf{O}_t, l, l_{\text{intent}}, z_{k,<j})
\end{align}
where $N_1$ and $N_2$ denote the lengths of the explicit and implicit action intent, respectively. Trained jointly on both domains, this phase drives the backbone to reason about manipulation in the shared intent space, thereby transferring human knowledge into the policy.

\textbf{Robot Domain Adaptation (Post-training).} After pre-training, the policy can plan in the shared intent space but cannot yet produce commands for a specific robot. 
% We therefore adapt it to the target embodiment: the backbone is fine-tuned on the robot dataset via LoRA, and a lightweight Action Decoder $\mathcal{D}_{act}$~\cite{univla} maps the implicit intent $\mathcal{Z}_{ic}$ to executable actions:
We therefore adapt it to the target embodiment. Specifically, we fine-tune the backbone on the robot dataset via LoRA and introduce a lightweight Action Decoder $\mathcal{D}_{act}$~\cite{univla} that decodes the implicit intent $\mathcal{Z}_{ic}$ into executable actions:
\begin{equation}
\hat{\mathbf{a}}_t, \dots, \hat{\mathbf{a}}_{t+m} = \mathcal{D}_{act}(\mathbf{a}_t, \dots, \mathbf{a}_{t+m} \mid \mathcal{Z}_{ic}, I_t, \mathbf{s}_t),
\end{equation}
where $\mathbf{s}_t$ denotes the proprioceptive state. The decoder is supervised by an L1 regression loss $\mathcal{L}_{action}$ between the predicted and ground-truth actions. The overall post-training objective combines this action term with the two intent losses:
\begin{equation}
\mathcal{L}_{posttrain} = \mathcal{L}_{explicit} + \mathcal{L}_{implicit} + \mathcal{L}_{action}.
\end{equation}
Through these two phases, our framework transfers semantic knowledge from human videos while ensuring precise kinematic control on the target robot.

\section{Experiment}
% In this section, we design our experiments to answer three core questions: (1) Can Action Intent function as a robust domain-invariant representation to transfer knowledge from human to robot domains? (2) Does our framework outperform existing baselines in both simulated and real-world manipulation tasks? (3) What additional advantages does our method yield beyond facilitating human-to-robot knowledge transfer?

In this section, we design our experiments to answer three core questions: (1) Can Action Intent function as a robust domain-invariant representation to transfer knowledge from human to robot domains? (2) Does our framework outperform existing baselines in both simulated and real-world manipulation tasks? (3) Does our method effectively disentangle action-relevant representations from scene-specific noise?
%2这里修改为我们的方法除了能得到人机迁移，还能带来其他什么效果
% (3) Does our method exhibit improved robustness and generalization to novel language instructions?

% \subsection{Baselines}
% 我们的basline包括transfer learning from human data(UniVLA\cite{univla}, LAPA\cite{lapa}), 使用 additional wrist-view camera inputs(MaIL\cite{jia2024mail}, MDT\cite{MDTreuss2024multimodal}), Transformer-based generalist cross robot embodiments policy(Octo\cite{team2024octo}) and the first open-source vla models(OpenVLA\cite{kim2024openvla}).
% \subsection{Baselines}
% Our baselines cover three key categories: (1) \textbf{Human-to-Robot Transfer} methods like UniVLA~\cite{univla} and LAPA~\cite{lapa}; (2) \textbf{Multi-View Policies} that incorporate additional wrist-view camera inputs (MaIL~\cite{jia2024mail}, MDT~\cite{MDTreuss2024multimodal}); and (3) \textbf{Generalist Models}, including the Transformer-based cross-embodiment policy Octo~\cite{team2024octo} and the first open-source VLA model OpenVLA~\cite{kim2024openvla}.

\subsection{Baselines}
We compare our method against a diverse set of baselines representing state-of-the-art approaches. These include methods leveraging transfer learning from human data, such as UniVLA~\cite{univla} and LAPA~\cite{lapa}. We also evaluate policies that utilize additional wrist-view camera inputs, specifically MaIL~\cite{jia2024mail} and MDT~\cite{MDTreuss2024multimodal}. Furthermore, we include generalist foundation models: Octo~\cite{team2024octo}, a Transformer-based cross-embodiment policy; OpenVLA~\cite{kim2024openvla}, a pioneering open-source Vision-Language-Action (VLA) model, and TriVLA~\cite{liu2025trivla}, a triple-system-based VLA model with episodic world modeling.

% \begin{itemize}
% \item UniVLA\cite{univla}: A cross-embodiment VLA framework that learns task-centric action representations from diverse videos.
% \item LAPA\cite{lapa}: Learns latent action priors from in-the-wild human videos via unsupervised learning.

% \item Octo\cite{team2024octo}: A Transformer-based generalist policy that unifies diverse robot embodiments through a shared action representation.

% \item MaIL\cite{jia2024mail}: Enhances imitation learning efficiency by integrating selective State Space Models for better scalability.

% \item MDT\cite{MDTreuss2024multimodal}: Leverages diffusion models to synthesize goal-conditioned action trajectories, handling multimodality effectively.

% \item OpenVLA\cite{kim2024openvla}: A VLA foundation model pretrained on the OpenX dataset, enabling generalization to new tasks.

% \end{itemize}

\subsection{Simulation Experiments}\label{sec:sim_exm}

{\bfseries Experimental Setup.} We conduct simulation experiments against baseline methods using the LIBERO benchmark~\cite{liu2023libero}. Specifically, we utilize manipulation tasks from the LIBERO-Spatial, LIBERO-Object, LIBERO-Goal, and LIBERO-10 suites. To facilitate adaptation, we employ our Rule-based Generation strategy to synthesize explicit action intents, thereby enriching these datasets with dense intent annotations. 
% Given that UniVLA\cite{univla} shares the most methodological similarities with our approach, we ensure a fair comparison by subjecting both models to an identical pre-training protocol on human (HOI4D\cite{hoi4d}, TACO\cite{taco}) and robot (Bridge-V2\cite{walke2023bridgedata}) datasets, followed by separate post-training (fine-tuning) on each LIBERO subset. During inference, we conduct 50 trials per task to evaluate performance.
Since UniVLA~\cite{univla} is the strongest human-to-robot transfer baseline, we adopt it as our primary competitor and ensure a strictly fair comparison. Both models are pre-trained under an identical protocol on human (HOI4D~\cite{hoi4d}, TACO~\cite{taco}) and robot (Bridge-V2~\cite{walke2023bridgedata}) datasets, and are then separately post-trained (fine-tuned) on each LIBERO subset. During inference, we conduct 50 trials per task to evaluate performance.

\begin{table}[h]
\centering
\renewcommand{\arraystretch}{1.3}
\setlength{\tabcolsep}{5pt}
\caption{Average Success Rate of our method and state-of-the-art baselines on the LIBERO benchmark across four evaluation suites.}
\label{table:simulation_libero}
\begin{tabular}{l|cccc|c}
\hline
Methods & Spatial & Object & Goal & Long & Avg. \\
\hline
LAPA~\cite{lapa} & 73.8 & 74.6 & 58.8 & 55.4 & 65.7 \\
Octo~\cite{team2024octo} & 78.9 & 85.7 & 84.6 & 51.1 & 75.1 \\
MDT~\cite{MDTreuss2024multimodal} & 78.5 & 87.5 & 73.5 & 64.8 & 76.1 \\
OpenVLA~\cite{kim2024openvla} & 84.7 & 88.4 & 79.2 & 53.7 & 76.5 \\
MaIL~\cite{jia2024mail} & 74.3 & 90.1 & 81.8 & 78.6 & 83.5 \\
UniVLA~\cite{univla} & 90.1 & 91.3 & 88.3 & 82.7 & 88.1 \\
TriVLA~\cite{liu2025trivla} &  91.2 & 93.8 & 89.8 &  73.2 & 87.0 \\
\hline
Ours & \textbf{93.4} & \textbf{94.2} & \textbf{91.2} & \textbf{86.4} & \textbf{91.3} \\
\hline
\end{tabular}
\end{table}

% \begin{table}[h]
% \centering
% \renewcommand{\arraystretch}{1.3} % 增加行高，使表格更易读，类似图片效果
% \begin{tabular}{l|cccc|c}
% \hline
% methods & Spatial & Object & Goal & Long & Avg. \\
% \hline
% LAPA & 73.8 & 74.6 & 58.8 & 55.4 & 65.7 \\
% %Diffusion Policy & 78.3 & 92.5 & 68.3 & 50.5 & 72.4 \\
% Octo & 78.9 & 85.7 & 84.6 & 51.1 & 75.1 \\
% MDT & 78.5 & 87.5 & 73.5 & 64.8 & 76.1 \\
% Openvla & 84.7 & 88.4 & 79.2 & 53.7 & 76.5 \\
% MaIL & 74.3 & 90.1 & 81.8 & 78.6 & 83.5 \\
% Univla & 90.1 & 91.3 & 88.3 & 82.7 & 88.1 \\
% \hline
% Ours & \textbf{93.4} & \textbf{94.2} & \textbf{91.2} & \textbf{86.4} & \textbf{91.3} \\
% \hline
% \end{tabular}
% \caption{Average Success Rate of our method and state-of-the-art baselines on the LIBERO benchmark across four evaluation suites.}
% \label{table:simulation_libero}
% \end{table}

% As shown in Tabel \ref{table:simulation_libero}, our methods outperforms Univla\cite{univla} in four bachmark, demonstrating 

% Our methods achieves the best performance compared to the strong baselines on all tasks. Remarkably, our
% 1、我们的方法性能好，在四个task上远超其余basline
% 2、尽管MaIL和MDT 利用了 additional wrist-view camera
% inputs，仍然没有我们的性能好
% 3、值得注意的是，我们的方法相较于LAPA提升了25.6\%的成功率，相较于Univla提升了3.2\%的成功率，证明了我们的action intent作为中间桥梁来讲human videos的知识源源不断迁移过来的的有效性
{\bfseries Results and Analysis.} 
As shown in Table~\ref{table:simulation_libero}, our method achieves state-of-the-art performance, significantly outperforming all baselines across the four evaluation suites (Spatial, Object, Goal, and Long). Specifically, our approach attains an impressive average success rate of 91.3\%.

Crucially, despite baselines such as MaIL~\cite{jia2024mail} and MDT~\cite{MDTreuss2024multimodal} benefiting from additional wrist-view camera inputs to capture fine-grained details, our method still surpasses them by margins of 7.8\% and 15.2\%, respectively. This demonstrates that explicit intent modeling can compensate for, or even exceed, the advantages typically gained from extra sensory modalities.

Furthermore, when comparing against human-to-robot transfer methods, our approach exhibits substantial gains: a 25.6\% improvement over LAPA and a 3.2\% boost over the strong competitor UniVLA. These results strongly validate the efficacy of Action Intent as a domain bridge. By explicitly modeling intent, our framework effectively unlocks the continuous transfer of manipulation knowledge from large-scale human videos, resulting in superior robustness and generalization. 

We further validate out-of-distribution generalization on the zero-shot CALVIN ABC$\rightarrow$D benchmark~\cite{calvin}, where our method again achieves state-of-the-art performance (4.02 average length, $+$5.8\% over UniVLA on the five-task sequence); see Appendix~\ref{appendix:calvin} for details.

\subsection{Real-World Experiments}\label{sec:real_exm}

% \begin{figure*}[ht]
%   \centering
%   \includegraphics[width=\linewidth]{picture/realworldexp.pdf}
%   \caption{(a)Real-world manipulation environment. The robot executes the picking, placing, and pushing tasks (T1–T12) listed in (b) within this cluttered tabletop scenario. (b)Success rates on the real-world benchmark: Univla vs. Ours across 12 tasks.}
%   \label{figure:realworld}
% \end{figure*}

\begin{figure}[t]
  \centering
  \includegraphics[width=\linewidth]{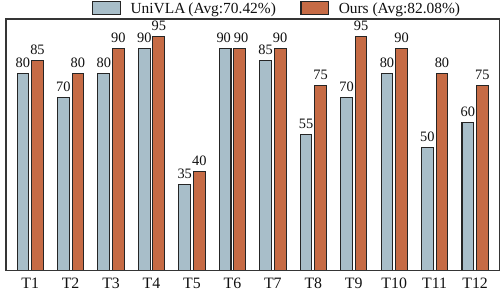}
  \caption{Success rates on the real-world benchmark across 12 tasks (T1--T12): UniVLA vs. Ours.}
  \label{figure:realworld}
\end{figure}

{\bfseries Experimental Setup.} We conduct real-world experiments utilizing the Astribot S1~\cite{astribotgao2025towards}. To isolate manipulation capabilities, we lock irrelevant degrees of freedom, activating only the 10-DoF right arm alongside its onboard RGB head camera.
To rigorously validate the efficacy of our approach, we curated a suite of 12 distinct tasks, collectively referred to as the Astribot Dataset (visualized in Appendix \ref{appendix:dataset}). These tasks are categorized into three hierarchical levels to probe different capabilities:

\begin{enumerate}
    \item \textbf{T1-T8: Standard Object Manipulation} (e.g., \textit{``Pick up the purple toy eggplant''}). These tasks evaluate basic visual grounding and grasping proficiency.
    \item \textbf{T9-T10: Spatial-Directional Awareness} (e.g., \textit{``Move the right lemon back''}). Designed to verify if our Action-Intent-Guided strategy effectively enhances spatial perception, enabling the robot to understand relative directional commands often found in human instructions.
    \item \textbf{T11-T12: Complex Spatial Reasoning} (e.g., \textit{``Place the left yellow lemon to the white platform in the right area''}). Building on T9-T10, these tasks require compositional reasoning over both object attributes and spatial relations.
\end{enumerate}

Crucially, the instructions in T9-T12 closely resemble the directive styles present in our human pre-training datasets (HOI4D~\cite{hoi4d} and TACO~\cite{taco}). Therefore, high performance on these tasks provides compelling evidence for the effective transfer of manipulation knowledge from the human video domain to the robot.

As UniVLA~\cite{univla} is the strongest human-to-robot transfer baseline, we adopt it as our primary competitor under an identical training protocol to ensure a strictly fair comparison.
% Given that UniVLA~\cite{univla} is the most competitive and methodologically similar baseline, we ensure a fair comparison by adopting an identical training protocol. 
Both models undergo pre-training on human (HOI4D~\cite{hoi4d}, TACO~\cite{taco}) and robot (Bridge-V2~\cite{walke2023bridgedata}) datasets, followed by fine-tuning on our Astribot Dataset. During inference, we conduct 20 trials per task.

{\bfseries Results and Analysis.} As illustrated in Fig.~\ref{figure:realworld}, our method demonstrates superior performance compared to the baseline across all 12 real-world tasks. Quantitatively, we attain an average success rate of 82.08\%, outperforming UniVLA (70.42\%) by a substantial margin of 11.66\%. In addition, our model runs at 8~Hz on a single RTX 4090, enabling real-time closed-loop control (about 5$\times$ faster than OpenVLA); a detailed efficiency analysis is provided in Appendix~\ref{appendix:efficiency}.

A detailed analysis highlights the distinct advantages of our method. On the Standard Object Manipulation tasks (T1-T8), our method achieves an 80.63\% success rate, compared to 73.13\% for UniVLA, reflecting a robust 7.5\% improvement in basic manipulation proficiency. Most critically, the performance gap widens significantly on the Spatial-Directional tasks (T9-T12), which demand the directional and relational reasoning that our human pre-training explicitly provides. Consequently, our method achieves an 85\% success rate, surpassing UniVLA (65\%) by a 20\% margin (further analyzed in Sec.~\ref{sec:human_to_robot}). This demonstrates that the proposed Action Intent serves as a robust, domain-invariant bridge, which not only enables efficient transfer of manipulation knowledge from human videos to the robot but also enhances the spatial and directional reasoning of the model.

% {\bfseries Generalization to Novel Language Instructions} %版面不够
% % \subsection{Generalization to Novel Language Instructions}
% To further assess the robustness of our policy against linguistic variations, we evaluated the models using novel language instructions that differ from the standard training templates in real-world deployment. As shown in Table~, we replaced explicit object identifiers (e.g., \textit{``purple toy eggplant''}) with more abstract, attribute-based descriptions (e.g., \textit{``the purple object''}). This modification challenges the model to ground visual attributes rather than relying on memorized object names.

% Quantitative results demonstrate that our method maintains superior performance even with these unseen instructions. While UniVLA experiences a performance drop, achieving a success rate of 55.6\% (10/18), our method attains a robust \textbf{72.2\% (13/18)}, surpassing the baseline by \textbf{16.6\%}. This improvement suggests that our Action Intent module facilitates deeper semantic grounding. By focusing on the underlying intent of the interaction rather than overfitting to specific lexical triggers, our framework exhibits stronger generalization to diverse linguistic formulations.

\subsection{Analysis of Human-to-Robot Knowledge Transfer}\label{sec:human_to_robot}

{\bfseries Experimental Setup.} To rigorously validate whether our method effectively transfers knowledge from human data, we evaluate along two dimensions: Action Intent prediction accuracy and real-world manipulation success rate.

We partition the Bridge dataset~\cite{walke2023bridgedata} at the trajectory level into a training set (95\%) and a test set (5\%). To isolate the contribution of human data, we instantiate four variants:
\begin{itemize}
    \item \textbf{Ours (Robot+Human)} \& \textbf{UniVLA (Robot+Human)}: pre-trained on the Bridge training set together with human datasets (TACO~\cite{taco}, HOI4D~\cite{hoi4d}).
    \item \textbf{Ours (Robot)} \& \textbf{UniVLA (Robot)}: pre-trained on the Bridge training set alone.
\end{itemize}

The evaluation proceeds in two stages. First, we evaluate whether our model predicts the correct Action Intent for a given task on the unseen Bridge test set. As the Action Intent comprises a directional movement (e.g., \textit{``move left''}) and a gripper state (e.g., \textit{``grasp''}), we report the classification accuracy of the two components separately. Second, to examine physical manipulation, we fine-tune all four variants on our Astribot Dataset and evaluate them on the Spatial-Directional and Complex Spatial Reasoning tasks (T9-T12), performing 20 trials per task. Together, these two stages reveal whether incorporating human data improves both intent understanding and downstream manipulation.

% \begin{table}[ht]
% \centering
% \renewcommand{\arraystretch}{1.3} % 增加行高，使表格更易读，类似图片效果
% \begin{tabular}{l|cc}
% \hline
% methods & Dir. Acc & Grip. Acc  \\
% \hline
% Ours (Robot) & 70.1 & 90.7\\
% Ours (Robot+Human) & 78.6 & 92.8 \\
% \hline
% \end{tabular}
% \caption{\textbf{Action Intent Prediction Analysis.} We report the Direction Accuracy (Dir. Acc) and Gripper State Accuracy (Grip. Acc) of our model variants on the unseen Bridge test set.}
% \label{table:analysis-human-action-intent}
% \end{table}

% {\bfseries Results and Analysis.}
% \textit{\textbf{(1)Action Intent Prediction.}} As shown in Table~\ref{table:analysis-human-action-intent}, incorporating human data yields improvements of \textbf{8.5\%} in Direction Accuracy and \textbf{2.1\%} in Gripper State Accuracy. These gains quantitatively validate that our method effectively transfers action intent from human data to enhance robot semantic understanding.

\begin{figure}[ht]
    \centering
    \includegraphics[width=\linewidth]{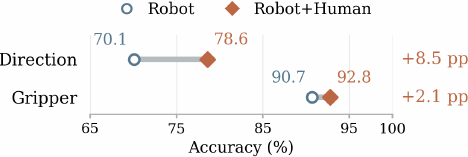}
    \caption{\textbf{Action Intent Prediction Analysis.}
    Direction and gripper state accuracy on the unseen Bridge test set
    with and without human training data.}
    \label{fig:analysis-human-action-intent}
\end{figure}

{\bfseries Results and Analysis.}
\textit{\textbf{(1) Action Intent Prediction.}}
As shown in Fig.~\ref{fig:analysis-human-action-intent},
incorporating human data improves direction accuracy from
70.1\% to 78.6\% and gripper state accuracy from 90.7\% to 92.8\%,
yielding gains of \textbf{8.5} and \textbf{2.1 percentage points},
respectively. These results support the effectiveness of
human-to-robot action intent transfer, with a larger improvement
in directional understanding.

\begin{table}[ht]
\centering
\renewcommand{\arraystretch}{1.3} % 增加行高，使表格更易读，类似图片效果
\setlength{\tabcolsep}{5pt}
\caption{\textbf{Success Rates on Real-World Tasks.} Comparison of UniVLA and our method across different pre-training data configurations (Robot vs. Robot+Human).}
\label{table:analysis-human-to-robot}
\begin{tabular}{l|cccc|c}
\hline
methods & T9 & T10 & T11 & T12 & Avg. \\
\hline
Univla (Robot) & 70 & 75 & 45 & 50 & 60.0 \\
Univla (Robot+Human) & 70 & 80 & 50 & 60 & 65.0 \\
\hline
Ours (Robot) & 75 & 85 & 70 & 65 & 73.8 \\
Ours (Robot+Human) & 95 & 90 & 80 & 75 & 85.0 \\
\hline
\end{tabular}
\end{table}

\textit{\textbf{(2)Real-world Manipulation Success Rate.}} Table~\ref{table:analysis-human-to-robot} presents the quantitative impact of human knowledge transfer on manipulation success rates. While both models benefit from human data, our method demonstrates superior data utilization. Specifically, incorporating human datasets increases our average success rate by 11.2\%, whereas Univla yields only a modest 5.0\% improvement under identical conditions. This widening performance gap serves as strong evidence that our method enables a more efficient and effective transfer of human knowledge compared to the baseline.

% \subsection{Generalization to Novel Language Instructions}
% %真机上 一个表，可合并到real-world
% \textcolor{red}{\subsection{Representation Disentanglement analyze}}
\subsection{Representation Disentanglement Analysis.}
\begin{figure}[t]
    \centering
    \includegraphics[width=\linewidth]
    {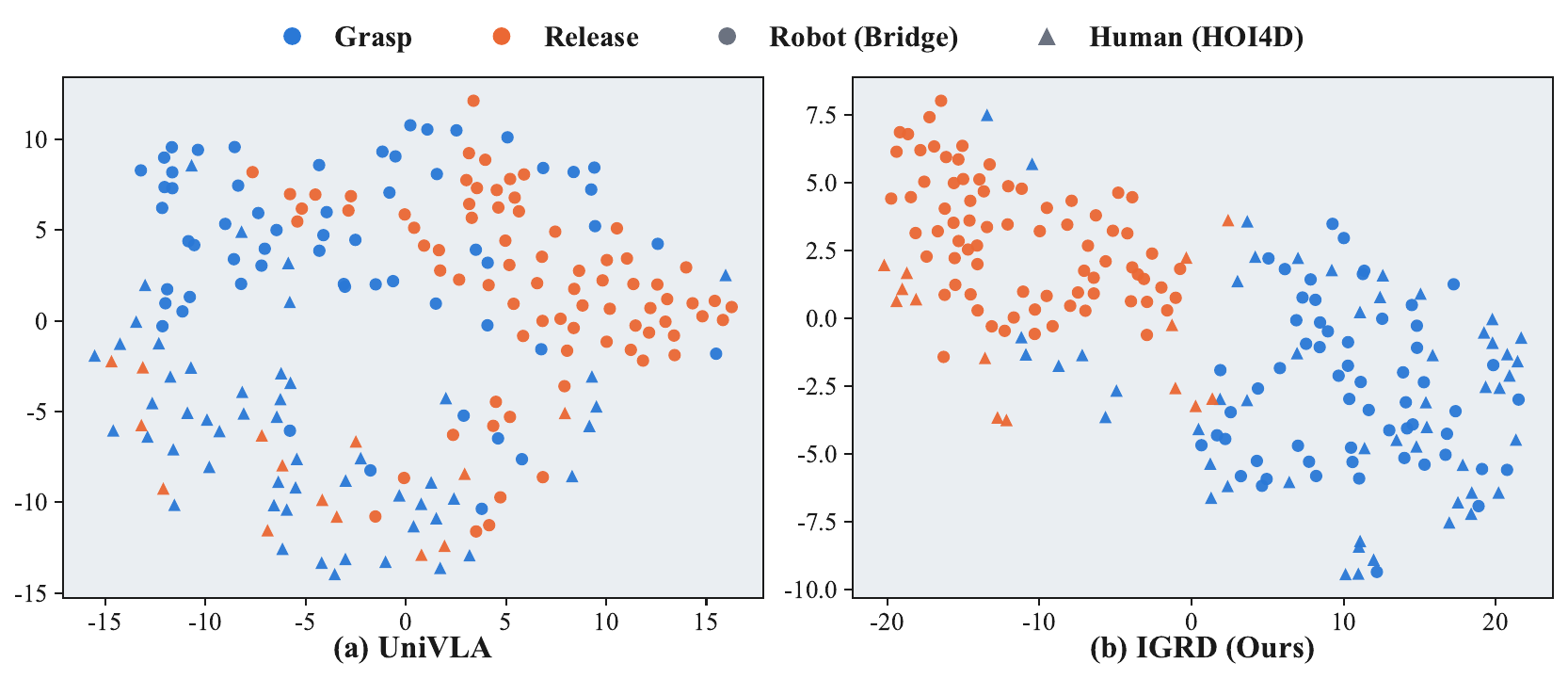}
    \caption{
    t-SNE visualization of latent action representations from
    (a) UniVLA and (b) IGRD (ours).
    Colors indicate action categories, and marker shapes denote
    human and robot domains.
    Our method yields clearer action separation and better
    cross-domain alignment within each action category.
    }
    \label{fig:representation_disentanglement}
\end{figure}

To examine whether our method learns action-relevant representations
shared across human and robot domains, we randomly sample a total
of 200 video clips covering grasp and release actions from HOI4D
and Bridge, extract their latent representations using UniVLA and
our method, and visualize them using t-SNE.
As shown in Fig.~\ref{fig:representation_disentanglement}(a),
UniVLA representations overlap across action categories while
remaining largely separated by domain, suggesting that action
semantics are entangled with domain-specific visual information.
In contrast, our method separates grasp from release while bringing
human and robot samples of the same action closer together
(Fig.~\ref{fig:representation_disentanglement}(b)).
This structure suggests that Action Intent captures shared action
semantics with reduced influence from scene-specific variation,
providing a consistent representation for human-to-robot
knowledge transfer.

\subsection{Ablation study}\label{sec:ablation}

{\bfseries Experimental Setup.}
To examine the contributions of explicit intent prediction,
decorrelation regularization, and intent-guided disentanglement,
we conduct a progressive ablation study in LIBERO.
All variants use the same human
(HOI4D~\cite{hoi4d}, TACO~\cite{taco}) and robot
(Bridge-V2~\cite{walke2023bridgedata}) pre-training datasets.
Starting from the full model, denoted as \textbf{M0 (Full)},
we successively remove these components.
First, \textbf{M1} removes explicit intent prediction during
policy pre-training and post-training: the backbone predicts
$\mathcal{Z}_{ic}$ directly from the observation and task
instruction, without first generating $l_{intent}$.
The intent-guided representation learner remains unchanged,
allowing this comparison to assess the contribution of the
intermediate explicit intent prediction step.
Second, \textbf{M2} additionally removes the cross-branch
decorrelation regularizer $\mathcal{L}_{\mathrm{dec}}$
while retaining the two-stage intent-guided disentanglement
procedure, isolating the contribution of explicit redundancy
reduction.
Finally, \textbf{M3} further removes intent-guided
disentanglement and instead uses reconstruction-based latent
action learning.
This final comparison assesses the importance of disentangling
action-relevant dynamics from scene-specific noise for
effective policy learning.
Implementation details are provided in
Appendix~\ref{app:progressive_ablation}.

\begin{table}[h]
\centering
\renewcommand{\arraystretch}{1.3}
\setlength{\tabcolsep}{4pt}
\caption{Progressive ablation study on LIBERO.
Starting from the full model (M0), M1--M3 successively
remove explicit intent prediction, decorrelation
regularization, and intent-guided disentanglement.
We report success rates (\%).}
\label{table:ablation_study}
\begin{tabular}{l|cccc|c}
\hline
Variant & Spatial & Object & Goal & Long & Avg. \\
\hline
M3 & 76.4 & 83.9 & 65.7 & 58.3 & 71.1 \\
M2 & 89.6 & 90.7 & 87.9 & 82.4 & 87.7 \\
M1 & 92.3 & 92.9 & 90.2 & 83.1 & 89.6 \\
M0 (Full) & \textbf{93.4} & \textbf{94.2} & \textbf{91.2} & \textbf{86.4} & \textbf{91.3} \\
\hline
\end{tabular}
\end{table}

{\bfseries Results and Analysis.}
Table~\ref{table:ablation_study} validates the effectiveness
of all three components.
Removing explicit intent prediction (M0 $\rightarrow$ M1)
reduces average success by 1.7\%,
with a larger drop of 3.3\% on LIBERO-Long,
supporting its role as an intermediate reasoning step
for long-horizon manipulation.
Further removing decorrelation regularization
(M1 $\rightarrow$ M2) decreases average success by 1.9\%,
demonstrating the benefit of explicitly reducing
cross-branch redundancy.
Replacing intent-guided disentanglement with
reconstruction-based latent action learning
(M2 $\rightarrow$ M3) causes the largest drop of 16.6 \%,
highlighting the importance of separating action-relevant
dynamics from scene-specific noise.
Together, these results support the complementary benefits
of explicit intent prediction, decorrelation regularization,
and intent-guided disentanglement for effective policy learning.

%1.8to do: 1、绘制两个图 2、完成所有experiment 抓住想表达的重点，借助gemini润色

\section{Conclusions}

In this work, we address human-to-robot knowledge transfer from the perspective of representation quality, and argue that effective transfer hinges on a clean action representation shared across humans and robots. We realize this through Intent-Guided Representation Disentanglement (IGRD), which represents manipulation as a shared action intent common to humans and robots. This intent bridges the two domains, transferring manipulation knowledge from human videos to the robot policy for execution. Extensive experiments in both simulation and the real world demonstrate that IGRD transfers human knowledge more effectively and achieves state-of-the-art performance. We hope this work motivates further study on how manipulation knowledge should be represented and transferred across embodiments.

\clearpage

\section*{LLM Usage Statement}

GPT-6 Astra was used to assist with implementation code generation, figure preparation, and language editing. Code generation was conducted under the authors’ direction and supervision throughout development. Framework diagrams were initially laid out with the tool and subsequently revised by the authors, while language editing was applied to author-written text. In addition, Gemini 2.5 Pro was used to generate task descriptions and explicit action-intent labels from human videos, as described in the annotation pipeline. The authors take full responsibility for the accuracy, originality, and integrity of the submitted work.

% \bigskip
% \noindent Thank you for reading these instructions carefully. We look forward to receiving your electronic files!

% References migrated to the CVPR template's main.bib.
{\small
\bibliographystyle{ieeenat_fullname}
\bibliography{main}
}

\clearpage

% Supplementary material begins on a separate page.
\appendix
\maketitlesupplementary

\section{Training Details}

% In this section, we present the training details of our DMS-VLA. We set the lenth of explicit action intent $N_1$ = 3 and implicit action intent $N_2$ = 4. All experiment are conducted in 8 H100. In Intent-Guided Latent Disentanglement stage, we perform 100,000 and 200,000 optimization step for stage A and stage B seperately. In Action Intents Policy Alignment stage we preform 20,000 optimization step for Joint pre-training stage. And conduct 100,000 optimization step for real-world deployment, 40,000 for Simulation Experiments in post-training.

% In this section, we present the implementation details of our AIT. We set the sequence lengths for explicit and implicit action intents to $N_1 = 3$ and $N_2 = 4$, respectively. All experiments are conducted on a cluster of 8 H100 GPUs. 
% Regarding the training schedule: In the Action-Intent-Guided Latent Disentanglement stage, we perform 100,000 optimization steps for Stage A and 200,000 steps for Stage B. In the Action Intent Policy Alignment stage, we execute 20,000 steps for joint pre-training. Subsequently, for the post-training phase, we conduct 100,000 steps for real-world deployment adaptation and 40,000 steps for simulation experiments.
In this section, we present the implementation details of our IGRD. We set the sequence lengths for explicit and implicit action intents to $N_1 = 3$ and $N_2 = 4$, respectively. Both codebooks $\mathcal{C}_s$ and $\mathcal{C}_{ic}$ contain 16 codewords, each with a dimensionality of 128. All experiments are conducted on a cluster of 8 H100 GPUs.
% In this section, we present the implementation details of our IGRD. We set the sequence lengths for explicit and implicit action intents to $N_1 = 3$ and $N_2 = 4$, respectively. All experiments are conducted on a cluster of 8 H100 GPUs.
Regarding the training schedule, in the Action-Intent-Guided Latent Disentanglement stage, we perform 100,000 optimization steps for Stage A and 200,000 steps for Stage B, with the decorrelation loss weight set to $\lambda_{\mathrm{dec}} = 10$ in Stage B. In the Action Intent Policy Alignment stage, we execute 20,000 steps for joint pre-training. Subsequently, for the post-training phase, we conduct 100,000 steps for real-world deployment adaptation and 40,000 steps for simulation experiments.

\begin{table}[htbp]
\centering
\small

\renewcommand{\arraystretch}{1.25}

\caption{Summary of the 12 tasks defined in the Astribot Dataset, categorized by required manipulation skills.}
\label{tab:astribot_tasks}
\begin{tabular}{@{} l c p{4.4cm} @{}}
\toprule
\textbf{Category} & \textbf{ID} & \textbf{Language Instruction} \\ 
\midrule

\multirow{8}{*}{\shortstack[l]{\textbf{Standard Object} \\ \textbf{Manipulation}}} 
 & T1 & Pick up the yellow lemon. \\
 & T2 & Pick up the white steamed bun. \\
 & T3 & Pick up the purple toy eggplant. \\
 & T4 & Pick up the orange toy orange. \\
 & T5 & Pick up the egg tart. \\
 & T6 & Pick up the white garlic. \\
 & T7 & Pick up the red lychee. \\
 & T8 & Pick up the small yellow rubber duck. \\ 
\midrule

\multirow{2}{*}{\shortstack[l]{\textbf{Spatial-Directional} \\ \textbf{Awareness}}} 
 & T9 & Move the right lemon back. \\
 & T10 & Push the brown kiwifruit back. \\ 
\midrule

\multirow{2}{*}{\shortstack[l]{\textbf{Complex Spatial} \\ \textbf{Reasoning}}} 
 & T11 & Place the yellow lemon to the white platform in the right area of the pegboard. \\
 & T12 & Place the green lemon to the black platform in the left area of the pegboard. \\ 
\bottomrule
\end{tabular}
\end{table}

\section{Prompts}
% In this section we provide the prompt used in VLM-Based Generation strategy.
% For generation Instruction for unlabeled human videos, We used the prompt:

In this section, we present the specific prompts employed in our VLM-based generation methodology.
\subsection{Instruction Generation for Unlabeled Human Videos}
\label{appendix:instruc_prompts}
To generate descriptive instructions for unlabeled human videos, we utilize the following prompt:

\begin{quote}
\small
\noindent \textbf{Role:} You are an expert VLA model trainer specializing in high-precision video annotation. \\

\noindent \textbf{Task:} Analyze the provided egocentric human manipulation video. Your goal is to generate dense, linguistically rich action labels to facilitate downstream VLA training. \\

\noindent \textbf{Annotation Guidelines:}
\begin{description}
    \item[Visual Granularity:] Describe objects with specific attributes, including color, texture, and material.
    \item[Spatial Awareness:] Clearly define the spatial relationship between the manipulator and the target objects.
    \item[Action Precision:] Use precise verbs (e.g., grasp, tilt, slide, insert).
    \item[Temporal Alignment:] Provide accurate start and end timestamps for each atomic action.
\end{description}

\noindent \textbf{Output Format:}
\begin{verbatim}
{
  "events": [
    {
      "id": 1,
      "event": "Pour the contents of 
      the white ceramic cup into the 
      green kettle on the right",
      "startTime": 0.00,
      "endTime": 1.39
    }
  ]
}
\end{verbatim}
\end{quote}

\subsection{Action Intent Inference for Human Videos}
\label{appendix:action_intent_prompts}
To infer the Action Intent from human actions, we use the following prompt:

\begin{quote}
\small
\noindent \textbf{Role:} You are an expert Motion Analyst specializing in egocentric video interpretation for VLA model training. \\

\noindent \textbf{Task:} Analyze the movement of the human right hand in the provided video. Your objective is to predict the hand's future trajectory and state by comparing the current frame with the state 1 second later. \\

\noindent \textbf{Annotation Guidelines:}
\begin{description}
    \item[Directional Trajectory:] Determine the dominant movement of the right hand within a 1-second look-ahead window. Choose the most accurate direction from the following action space: [Move Right, Move Left, Move Front, Move Back, Move Up, Move Down].
    \item[Grasping State:] Identify the hand's interaction state 1 second into the future. Label it as either [Grasp] or [Release].
    \item[Temporal Precision:] Map these future-oriented labels to the corresponding startTime and endTime of the current observation.
\end{description}

\noindent \textbf{Output Format:}
\begin{verbatim}
{
  "events": [
    {
      "id": 1,
      "event": "Move Up, Grasp",
      "startTime": 0.20,
      "endTime": 0.30
    }
  ]
}
\end{verbatim}
\end{quote}

\section{Reliability of VLM-Generated Pseudo-Labels}
\label{appendix:reliability}

We provide a quantitative analysis of the VLM-generated pseudo-labels used for our human videos from two complementary perspectives: the \emph{reliability} of the VLM annotations, and the \emph{output robustness} of the trained model.

\paragraph{Reliability of VLM Annotations.}
To quantify the intrinsic reliability of the VLM annotations prior to human correction, we randomly sample the VLM outputs on the human videos and ask professional annotators to verify their correctness. As reported in Table~\ref{tab:reliability_input}, 96.7\% (290/300) of the language instructions are accurate and contextually reasonable. For the explicit action intents, the accuracy reaches 96.4\% (482/500) for ``grasp'' and 82.4\% (412/500) for ``move''. In contrast, the intents for the robot dataset are derived directly from precise end-effector coordinates via our rule-based strategy, providing a 100\% ground-truth baseline. Guided by this audit, the remaining erroneous annotations are corrected by human annotators, yielding a clean training set.

\paragraph{Output Robustness and Motion Ambiguity.}
% We further examine the intent prediction of the trained model on 10{,}000 unseen Bridge trajectories. 
We further evaluate intent prediction on 10{,}000 unseen Bridge trajectories using the same Robot+Human model reported in Figure ~\ref{fig:analysis-human-action-intent}.
As shown in Table~\ref{tab:reliability_output}, under strict coordinate-based matching the model attains 92.8\% accuracy for ``grasp'' and 78.6\% for ``move''. Importantly, the comparatively lower ``move'' number does not indicate model failure, but rather reflects the inherent multi-axis ambiguity of natural motion under a strict single-axis metric. For instance, reaching an object at the top-left involves both upward and leftward components, so predicting either is semantically valid even when it mismatches the dominant coordinate axis. To account for this, human evaluators re-assessed 200 predictions for semantic correctness, and the accuracy rises to 85.5\% (171/200) for ``move'' and 94.0\% (188/200) for ``grasp''.

Together, these results confirm that the VLM provides highly reliable supervision, and that the residual ``move'' mismatch reflects natural motion ambiguity rather than genuine prediction errors.

\begin{table}[htbp]
\centering
\small
\renewcommand{\arraystretch}{1.25}
\caption{Accuracy of the VLM pseudo-labels prior to human correction. Human-video labels are audited on random samples, whereas robot labels are derived from precise kinematic coordinates.}
\label{tab:reliability_input}
\begin{tabular}{l c c}
\toprule
\textbf{Label Type} & \textbf{\# Samples} & \textbf{Accuracy} \\
\midrule
Language Instruction (human) & 300 & 96.7\% \\
Explicit Intent -- Grasp (human) & 500 & 96.4\% \\
Explicit Intent -- Move (human) & 500 & 82.4\% \\
Explicit Intent (robot, rule-based) & -- & 100\% \\
\bottomrule
\end{tabular}
\end{table}

\begin{table}[htbp]
\centering
\small
\setlength{\tabcolsep}{4pt}
\renewcommand{\arraystretch}{1.25}
\caption{Intent prediction accuracy of the trained model on the unseen Bridge test set. The gap on ``move'' reflects multi-axis motion ambiguity under a strict single-axis metric, as confirmed by human semantic evaluation.}
\label{tab:reliability_output}
\begin{tabular}{l c c}
\toprule
\textbf{Evaluation Protocol} & \textbf{Move Acc.} & \textbf{Grasp Acc.} \\
\midrule
Strict coordinate matching (10{,}000) & 78.6\% & 92.8\% \\
Human semantic evaluation (200) & 85.5\% & 94.0\% \\
\bottomrule
\end{tabular}
\end{table}

\section{Out-of-Distribution Generalization on CALVIN}
\label{appendix:calvin}

\textbf{Experimental Setup.} To further probe the generalization of our framework, we evaluate on the CALVIN benchmark~\cite{calvin}, which comprises 34 tasks with free-form language instructions spanning a wide spectrum of skills, ranging from basic pick-and-place operations to articulated object manipulation. The benchmark contains four distinct environments (A--D), each featuring a Franka Panda arm for tabletop manipulation. We adopt the most challenging \emph{ABC$\rightarrow$D} setting, in which policies are trained on demonstrations from environments A, B, and C, followed by zero-shot evaluation in the held-out environment D. The evaluation protocol consists of 1{,}000 unique instruction chains, each comprising five consecutive tasks, thereby rigorously assessing long-horizon generalization. Consistent with our LIBERO protocol and for a fair comparison, our model takes only third-view RGB images and language instructions as input, and is fine-tuned on the CALVIN ABC demonstrations.

\textbf{Results.} As reported in Table~\ref{tab:calvin_results_supp}, our method establishes a new state of the art on CALVIN ABC$\rightarrow$D. It completes all five tasks in sequence with a 62.3\% success rate, surpassing the strongest baseline UniVLA (56.5\%) by 5.8\% and OpenVLA by 18.8\%. Correspondingly, the average number of consecutively completed tasks increases to 4.02, the highest among all compared methods. Moreover, our advantage over UniVLA is more pronounced on longer sequences, growing from 1.7\% on the first task to 5.8\% on the fifth. 
% This trend indicates that distilling manipulation semantics from human videos is particularly beneficial for complex, long-horizon reasoning, corroborating the conclusions drawn from our LIBERO and real-world experiments.
This trend indicates that transferring manipulation knowledge from human videos is particularly beneficial for complex, long-horizon reasoning, corroborating the conclusions drawn from our LIBERO and real-world experiments.

% \begin{table}[htbp]
%     \centering
%     \caption{\textbf{Language-conditioned visuomotor control on CALVIN ABC$\rightarrow$D.} We report the success rate of completing $1$--$5$ tasks in a row and the average number of completed tasks (Avg. Len.) per instruction chain. All methods use only third-view RGB and language inputs. $^{*}$Reproduced with action-chunk prediction.}
%     \label{tab:calvin_results_supp}
%     \renewcommand{\arraystretch}{1.15}
%     \begin{tabular}{l|ccccc|c}
%         \toprule
%         \multirow{2}{*}{Method} & \multicolumn{5}{c|}{Task completed in a row (\%) $\uparrow$} & Avg. \\
%          & 1 & 2 & 3 & 4 & 5 &  Len.\\
%         \midrule
%         RT-1~\cite{brohan2022rt}          & 53.3 & 22.2 & 9.4  & 3.8  & 1.3  & 0.90 \\
%         RoboFlamingo~\cite{roboflamingo}  & 82.4 & 61.9 & 46.6 & 33.1 & 23.5 & 2.48 \\
%         SuSIE~\cite{susie}                & 87.0 & 69.0 & 49.0 & 38.0 & 26.0 & 2.69 \\
%         GR-1~\cite{gr1}                   & 85.4 & 71.2 & 59.6 & 49.7 & 40.1 & 3.06 \\
%         OpenVLA$^{*}$~\cite{kim2024openvla} & 91.3 & 77.8 & 62.0 & 52.1 & 43.5 & 3.27 \\
%         CLOVER~\cite{clover}              & 96.0 & 83.5 & 70.8 & 57.5 & 45.4 & 3.53 \\
%         RoboDual~\cite{robodual}          & 94.4 & 82.7 & 72.1 & 62.4 & 54.4 & 3.66 \\
%         UniVLA~\cite{univla}              & 95.5 & 85.8 & 75.4 & 66.9 & 56.5 & 3.80 \\
%         \midrule
%         Ours                              & \textbf{97.2} & \textbf{89.1} & \textbf{81.4} & \textbf{71.7} & \textbf{62.3} & \textbf{4.02} \\
%         \bottomrule
%     \end{tabular}
% \end{table}
\begin{table}[htbp]
    \centering
    \caption{\textbf{Language-conditioned visuomotor control on CALVIN ABC$\rightarrow$D.} We report the success rate of completing $1$--$5$ tasks in a row and the average number of completed tasks (Avg. Len.) per instruction chain. All methods use only third-view RGB and language inputs. $^{*}$Reproduced with action-chunk prediction.}
    \label{tab:calvin_results_supp}
    \small
    \setlength{\tabcolsep}{5.5pt}
    \renewcommand{\arraystretch}{1.15}
    \begin{tabular}{@{}l|ccccc|c@{}}
        \toprule
        \multirow{2}{*}{Method}
        & \multicolumn{5}{c|}{Tasks in a row (\%) $\uparrow$}
        & Avg. \\
        & 1 & 2 & 3 & 4 & 5 & Len. \\
        \midrule
        RT-1~\cite{brohan2022rt}
        & 53.3 & 22.2 & 9.4 & 3.8 & 1.3 & 0.90 \\
        RoboFlamingo~\cite{roboflamingo}
        & 82.4 & 61.9 & 46.6 & 33.1 & 23.5 & 2.48 \\
        SuSIE~\cite{susie}
        & 87.0 & 69.0 & 49.0 & 38.0 & 26.0 & 2.69 \\
        GR-1~\cite{gr1}
        & 85.4 & 71.2 & 59.6 & 49.7 & 40.1 & 3.06 \\
        OpenVLA$^{*}$~\cite{kim2024openvla}
        & 91.3 & 77.8 & 62.0 & 52.1 & 43.5 & 3.27 \\
        CLOVER~\cite{clover}
        & 96.0 & 83.5 & 70.8 & 57.5 & 45.4 & 3.53 \\
        RoboDual~\cite{robodual}
        & 94.4 & 82.7 & 72.1 & 62.4 & 54.4 & 3.66 \\
        UniVLA~\cite{univla}
        & 95.5 & 85.8 & 75.4 & 66.9 & 56.5 & 3.80 \\
        \midrule
        Ours
        & \textbf{97.2} & \textbf{89.1} & \textbf{81.4}
        & \textbf{71.7} & \textbf{62.3} & \textbf{4.02} \\
        \bottomrule
    \end{tabular}
\end{table}

\section{Implementation Details of Progressive Ablations}
\label{app:progressive_ablation}

We construct four progressively simplified variants,
denoted as M0--M3, to evaluate explicit intent prediction,
cross-branch decorrelation, and intent-guided disentanglement.
The modifications to representation learning and policy
training are detailed below.

\paragraph{M0: Full Model.}
The full model learns $\mathcal{Z}_{ic}$ through the two-stage
intent-guided disentanglement procedure.
Stage A learns scene representations conditioned on the task
instruction $l$ and explicit intent $l_{intent}$.
Stage B learns implicit intent representations with the scene
codebook frozen and includes the cross-branch decorrelation
regularizer $\mathcal{L}_{\mathrm{dec}}$.
During policy pre-training and post-training, the backbone
first predicts $l_{intent}$ and then generates
$\mathcal{Z}_{ic}$.

\paragraph{M1: Removing Explicit Intent Prediction.}
This variant changes only policy training and inference.
The representation learner, including both disentanglement
stages and the decorrelation regularizer, is identical to
that of M0.
The policy directly predicts $\mathcal{Z}_{ic}$ conditioned
on the observation and task instruction, without generating
$l_{intent}$.
Accordingly, we remove $\mathcal{L}_{\mathrm{explicit}}$
from both policy training objectives:
\begin{equation}
\begin{aligned}
\mathcal{L}_{\mathrm{pretrain}}^{\mathrm{M1}}
&= \mathcal{L}_{\mathrm{implicit}}, \\
\mathcal{L}_{\mathrm{posttrain}}^{\mathrm{M1}}
&= \mathcal{L}_{\mathrm{implicit}}
 + \mathcal{L}_{\mathrm{action}}.
\end{aligned}
\end{equation}
Explicit intent annotations are still used to train the
representation learner in Stage A.

\paragraph{M2: Removing Decorrelation Regularization.}
Building on M1, this variant sets
$\lambda_{\mathrm{dec}}=0$ during Stage B.
Stage A, the initialization of Stage B from Stage A,
and the frozen scene codebook are retained.
Stage B is therefore optimized using only the reconstruction
and vector-quantization objective in Eq.~\ref{eq:vqloss}.
Policy training follows the same direct prediction procedure
and objectives as M1.
This comparison isolates the contribution of the
decorrelation regularizer within the intent-guided
disentanglement framework.

\paragraph{M3: Removing Intent-Guided Disentanglement.}
Building on M2, this variant replaces the entire two-stage
intent-guided disentanglement procedure with a single-stage
latent action learner.
Given two frames $o_t$ and $o_{t+k}$, we retain DINOv2 as
the visual encoder $f_{\mathrm{DINO}}$:
\begin{equation}
I_t = f_{\mathrm{DINO}}(o_t), \qquad
I_{t+k} = f_{\mathrm{DINO}}(o_{t+k}).
\end{equation}
An encoder $E_{\mathrm{lat}}$ processes the paired visual
features, and a single latent action codebook
$\mathcal{C}_{\mathrm{lat}}$ quantizes its output:
\begin{equation}
\mathcal{Z}_{\mathrm{lat}}
=
Q_{\mathcal{C}_{\mathrm{lat}}}
\bigl(E_{\mathrm{lat}}(I_t,I_{t+k})\bigr),
\end{equation}
where $Q$ denotes vector quantization.
A feature-wise decoder reconstructs the future-frame
features from the current-frame features and the latent
action tokens:
\begin{equation}
\widehat{I}_{t+k}
=
D_{\mathrm{lat}}(I_t,\mathcal{Z}_{\mathrm{lat}}).
\end{equation}
The encoder, codebook, and decoder are trained jointly
in a single stage using
\begin{equation}
\mathcal{L}_{\mathrm{lat}}
=
\left\|\widehat{I}_{t+k}-I_{t+k}\right\|_2^2
+
\mathcal{L}_{\mathrm{VQ}}.
\end{equation}
Reconstruction is performed in the DINOv2 feature space.
The representation learner uses neither task-text nor
explicit-intent conditioning and contains no separate scene
branch, intent-guided initialization, or cross-branch
decorrelation regularizer.
Its latent tokens encode visual transitions without an
explicit mechanism for separating intent-relevant motion
from scene variation.

For policy learning, $\mathcal{Z}_{\mathrm{lat}}$ replaces
$\mathcal{Z}_{ic}$ as the prediction target and the
representation passed to the action decoder.
As in M1 and M2, the policy receives the observation and
task instruction and predicts the latent tokens without
generating $l_{intent}$.
The policy objectives retain latent-token prediction and,
during post-training, action supervision.
The comparison between M2 and M3 assesses the importance
of disentangling action-relevant dynamics from
scene-specific noise for effective policy learning.

\section{Inference Efficiency}
\label{appendix:efficiency}

We analyze the real-world inference efficiency of our model on a single RTX 4090 GPU. Our model achieves a control frequency of 8~Hz, which is sufficient for smooth real-time closed-loop manipulation. Compared with OpenVLA~\cite{kim2024openvla}, which autoregressively decodes low-level action tokens and runs at only 1.5~Hz, our model is roughly 5$\times$ faster, as we directly decode the implicit action intent $\mathcal{Z}_{ic}$ into actions rather than generating actions token by token. Compared with UniVLA~\cite{univla} (10~Hz), the intermediate intent generation introduces a modest 2~Hz overhead. Nevertheless, this overhead is well justified: our method attains a substantially higher real-world success rate (82.08\% vs. 70.42\%), and the improved reliability reduces corrective retries during execution. Overall, our framework offers a favorable accuracy--efficiency trade-off for practical deployment.

\begin{table}[htbp]
\centering
\small
\renewcommand{\arraystretch}{1.25}
\caption{Real-world control frequency on a single RTX 4090.}
\label{tab:efficiency}
\begin{tabular}{l c c c}
\toprule
\textbf{Method} & OpenVLA & UniVLA & Ours \\
\midrule
\textbf{Control Freq. (Hz)} & 1.5 & 10 & 8 \\
\bottomrule
\end{tabular}
\end{table}

\section{Astribot dataset}
\label{appendix:dataset}
% In this section we present our carefully designed 12 distinct tasks (Astribot dataset) in tabel

% T1：pick up yellow lemon
% T2：pick up white steamed bun
% T3：pick up purple toy eggplant
% T4：pick up orange toy orange
% T5：pick up egg tart
% T6：pick up white garlic
% T7：pick up red lychee
% T8：pick up small yellow rubber duck
% T9：Move the right lemon back
% T10：Push the brown kiwifruit back
% T11：place the yellow lemon to the white platform in the right area of the pegboard
% T12：place the green lemon to the black platform in the left area of the pegboard

% In this section, we provide a comprehensive list of the 12 distinct manipulation tasks defined in the Astribot dataset, as presented in Table~\ref{tab:astribot_tasks}. These tasks are designed to evaluate the model's ability to ground language instructions into precise physical actions across various objects and spatial configurations.

In this section, we provide a comprehensive list of the 12 distinct manipulation tasks constituting the Astribot dataset, as summarized in Table~\ref{tab:astribot_tasks}. These tasks are curated to evaluate our model's capability to ground natural language instructions into precise physical actions across diverse objects and spatial configurations.

% \section{Qualitative Results}
% In this section, we provide the Qualitative Results in both real-world and simulation environments as shown in Fig\ref{figure:qr_real} and Fig.\ref{figure:qe_sim} our compared with univla\cite{univla}.

\section{Qualitative Results}
\begin{figure}[htbp]
  \centering
  \includegraphics[width=\linewidth]{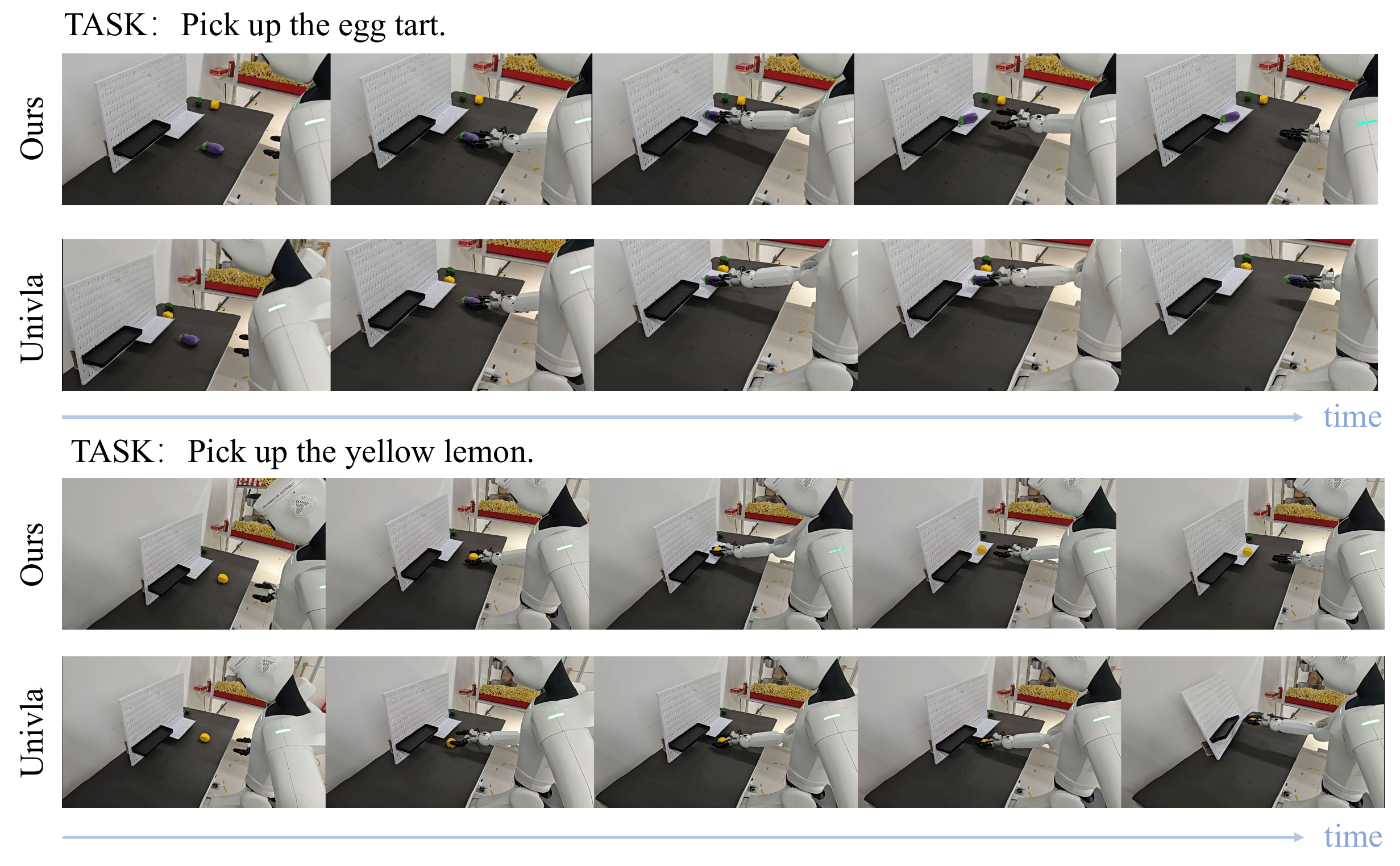}
  \caption{Qualitative Results in real-world environments.}
  \label{figure:qr_real}
\end{figure}

\begin{figure}[htbp]
  \centering
  \includegraphics[width=\linewidth]{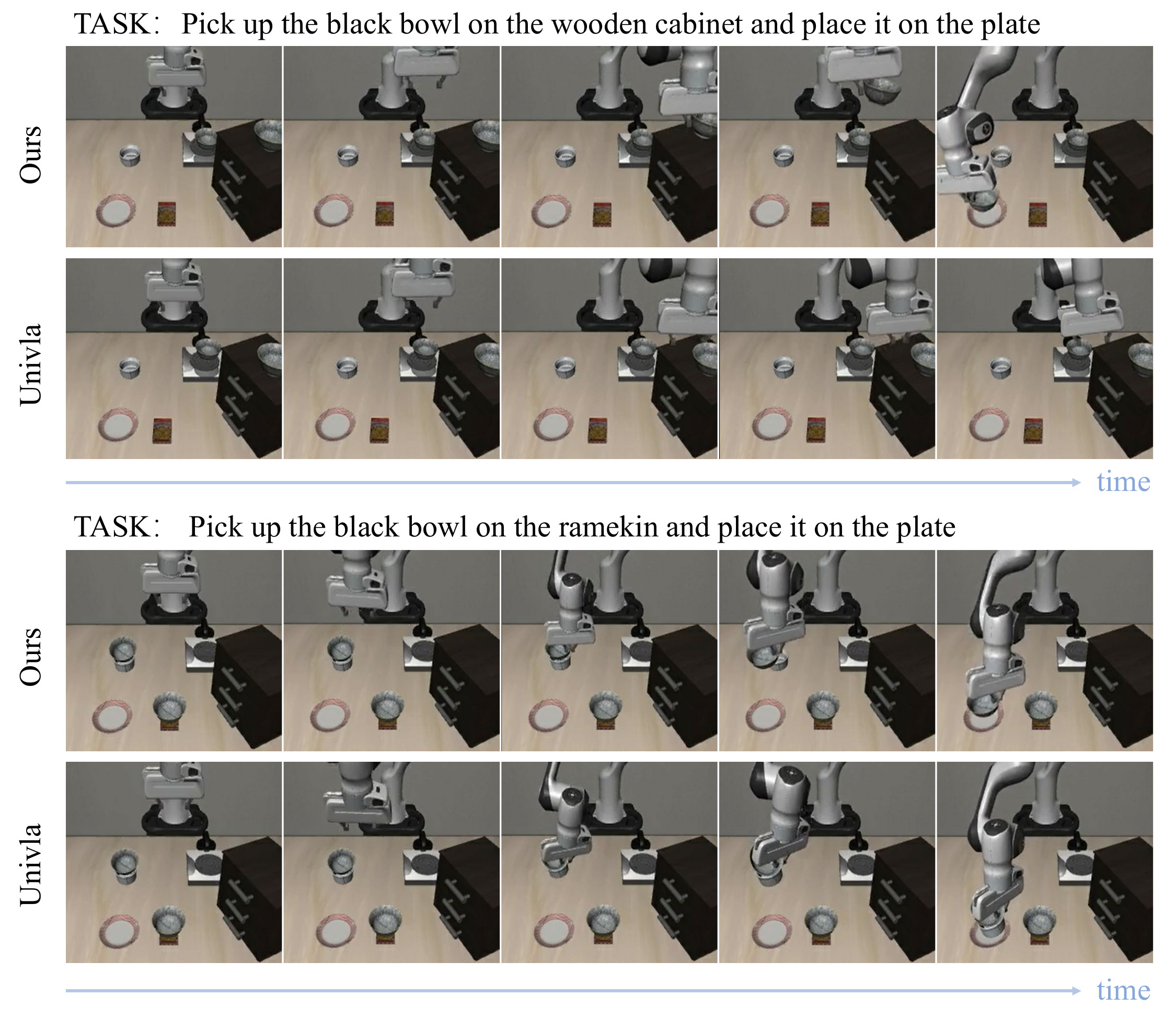}
  \caption{Qualitative Results in simulation environments.}
  \label{figure:qe_sim}
\end{figure}

In this section, we present qualitative comparisons between our method and UniVLA~\cite{univla} across both real-world and simulation environments. Detailed visual results are illustrated in Fig.~\ref{figure:qr_real} and Fig.~\ref{figure:qe_sim}, respectively.

\section{Sensitivity to Explicit Intent Vocabulary}
\label{sec:intent_vocabulary}

\paragraph{Explicit intents and continuous actions.}
The explicit intent vocabulary provides coarse semantic guidance for representation learning rather than a complete parameterization of robot actions. The six directional primitives describe translational tendencies, while \texttt{Grasp} and \texttt{Release} describe gripper changes. In particular, the dominant axis of the displacement between two frames summarizes net translation rather than the full intervening path.

Our policy is not restricted to executing these discrete primitives. During latent disentanglement, the model learns from visual transitions through a reconstruction objective, allowing implicit intent tokens $\mathcal{Z}_{ic}$ to encode motion information beyond the explicit labels. During policy alignment, the action decoder conditions on $\mathcal{Z}_{ic}$, visual observations, and proprioceptive states, and is supervised using the original continuous robot actions. Thus, motion components omitted from the explicit vocabulary, such as rotation, remain available through visual transitions and continuous-action supervision. This design provides a route for learning such components without requiring a separate explicit label for each one. To examine the robustness of our method to the granularity of explicit intent supervision, we expand the vocabulary with rotational primitives and evaluate its effects on reconstruction during latent disentanglement and downstream manipulation performance.

\paragraph{Experimental setup.}
% We compare the default vocabulary $\mathrm{B}$ with an expanded vocabulary $\mathrm{B}^{+}$:
% \begin{equation}
% \begin{aligned}
% \mathcal{A}_{\mathrm{B}} = \{\,&\texttt{Move Right},\texttt{Move Left},\\
% &\texttt{Move Front},\texttt{Move Back},\\
% &\texttt{Move Up},\texttt{Move Down},\\
% &\texttt{Grasp},\texttt{Release}\,\},\\
% \mathcal{A}_{\mathrm{B}^{+}} =\;&\mathcal{A}_{\mathrm{B}}\cup
% \{\texttt{Rotate Left},\texttt{Rotate Right}\}.
% \end{aligned}
% \end{equation}
We compare the default vocabulary $\mathrm{B}$ with an expanded vocabulary $\mathrm{B}^{+}$:
\begin{equation}
\begin{aligned}
\mathcal{A}_{\mathrm{B}} = \{\,
&\text{Move Right/Left/Front/Back/Up/Down},\\
&\text{Grasp/Release}\,\},\\
\mathcal{A}_{\mathrm{B}^{+}} =\;&\mathcal{A}_{\mathrm{B}}\cup
\{\text{Rotate Left},\text{Rotate Right}\}.
\end{aligned}
\end{equation}
For the expanded vocabulary, we repeat Action-Intent-Guided Latent Disentanglement and Action Intent Policy Alignment, including joint pre-training and robot post-training, following the original experimental setting. We evaluate both variants on the same 12 real-world tasks, with 20 trials per task and 240 trials per variant.

\begin{figure}[t]
    \centering
    \includegraphics[width=\linewidth]{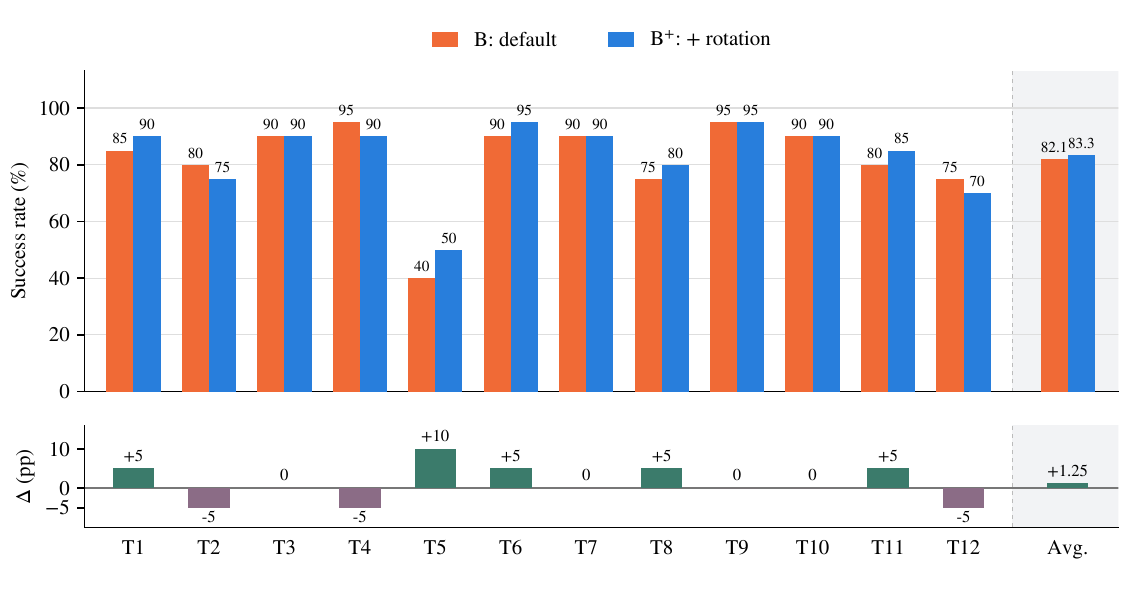}
    \caption{\textbf{Effect of explicit intent vocabulary on real-world performance.} Top: success rates for the default vocabulary $\mathrm{B}$ and the expanded vocabulary $\mathrm{B}^{+}$, using 20 trials per task. Avg. is the unweighted mean over T1--T12. Bottom: task-wise changes in percentage points, computed as $\mathrm{B}^{+}-\mathrm{B}$. A 5-point change on an individual task corresponds to one successful trial.}
    \label{fig:intent_vocabulary_realworld}
\end{figure}

\paragraph{Real-world performance.}
As shown in Fig.~\ref{fig:intent_vocabulary_realworld}, adding rotational primitives increases the average success rate from 82.08\% to 83.33\%, a gain of 1.25 percentage points computed from the unrounded means. The largest observed gain is 10 percentage points on T5; all other task-wise changes are within $\pm5$ percentage points. These results show a modest aggregate benefit from the tested vocabulary extension, suggesting limited sensitivity to this choice on our real-world benchmark. They do not establish statistical equivalence between the two variants or quantify disagreement between rule-derived labels and independently annotated motion primitives.

\begin{figure}[t]
    \centering
    \includegraphics[width=\linewidth]{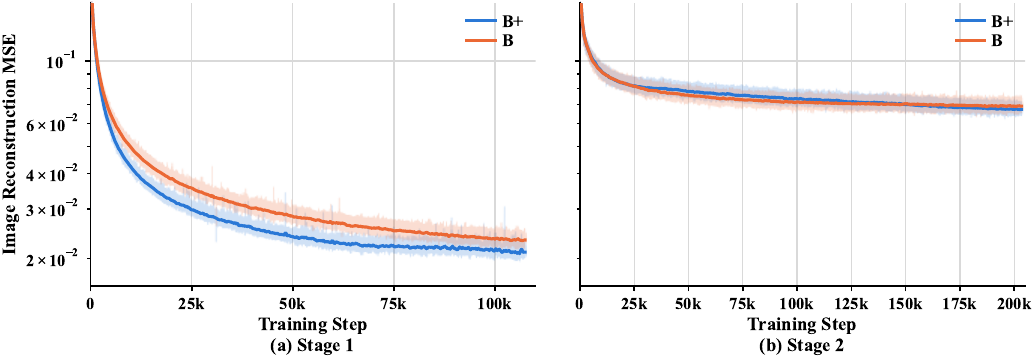}
    \caption{\textbf{Reconstruction training curves under different intent vocabularies.} The Stage 1 and Stage 2 plots correspond to Stage A, scene learning, and Stage B, intent disentanglement, respectively. With $\mathrm{B}^{+}$, Stage A reaches lower reconstruction errors earlier and ends at a lower MSE. In Stage B, $\mathrm{B}^{+}$ has slightly higher errors over much of training but a slightly lower final MSE.}
    \label{fig:intent_vocabulary_reconstruction}
\end{figure}

\paragraph{Effect on latent disentanglement.}
% Figure~\ref{fig:intent_vocabulary_reconstruction} shows that the expanded vocabulary changes the optimization behavior of both stages. In Stage A, richer intent conditioning provides additional motion information for reconstruction, consistent with the earlier reduction and lower final value of the reconstruction loss. In Stage B, where explicit text conditioning is replaced by implicit intent tokens and the scene codebook is frozen, the expanded-vocabulary variant exhibits slightly higher reconstruction errors over much of training before reaching a slightly lower final error. 

% Taken together, finer explicit intents improve the final reconstruction error while yielding a modest observed gain in downstream success. This supports the use of the default compact vocabulary on the evaluated tasks, with potential benefits from additional motion labels. The results are consistent with continuous-action supervision complementing coarse intent guidance, but do not isolate whether or how subsequent training compensates for omitted intent categories.
Figure~\ref{fig:intent_vocabulary_reconstruction} illustrates the effect of finer-grained intent guidance on reconstruction during both stages. In Stage A, the expanded vocabulary provides richer motion information, leading to a faster decrease in reconstruction loss and a lower final error. In Stage B, with the scene codebook frozen and explicit text conditioning replaced by implicit intent tokens, the expanded-vocabulary variant exhibits slightly higher reconstruction errors over much of training but ultimately reaches a slightly lower final error.

Overall, finer-grained intent guidance improves reconstruction, while increasing downstream success from 82.08\% to 83.33\%. This modest gain suggests that our method is relatively robust to the tested expansion of the explicit intent vocabulary. The default compact vocabulary therefore provides effective guidance for the evaluated tasks, with finer motion labels offering incremental benefits.

\section{Comparison with CLAP on LIBERO}\label{appendix:clap}

To further validate our method against the latest advances, we compare IGRD with CLAP~\cite{zhang2026clap}, a recently proposed approach that also learns VLA policies from human videos. CLAP contrastively aligns a visual latent-dynamics space with a discrete action codebook learned from robot trajectories, so that action-free human videos can be supervised with pseudo action tokens. Since CLAP was originally pre-trained on different data, we re-implement it under exactly the same protocol as UniVLA and our method (Sec.~\ref{sec:sim_exm}), so that the comparison reflects the learning paradigm rather than the training data.

\paragraph{Reproduction details.}
We follow the original four-stage pipeline of CLAP and restrict all training data to the datasets used in our experiments:
\begin{itemize}
    \item \textbf{Stage A (Act-VAE).} Using only the action trajectories of Bridge-V2~\cite{walke2023bridgedata}, we train the action encoder, the action codebook, and the action decoder. This yields a robot action tokenizer.
    \item \textbf{Stage B (VD-VAE).} On Bridge-V2, HOI4D~\cite{hoi4d}, and TACO~\cite{taco}, we freeze the visual backbone together with the Stage-A action encoder and codebook, and train the visual-dynamics module and the environment codebook. The resulting visual tokenizer maps frame transitions of action-free human videos to tokens in the Stage-A action vocabulary.
    \item \textbf{Stage C (CLAP-NTP pre-training).} With both tokenizers frozen, we train the VLA by next-token prediction on the same robot--human data mixture. Robot samples are supervised with ground-truth action tokens; human samples are supervised with the visual pseudo action tokens produced in Stage B.
    \item \textbf{Stage D (per-suite post-training).} For each of LIBERO-Spatial, -Object, -Goal, and -Long, we initialize independently from the Stage-C checkpoint. We first perform next-token-prediction adaptation on the target suite and then apply CLAP's RF+KM post-training. This yields four independent policies, matching the per-suite fine-tuning of the other methods.
\end{itemize}
For human videos, CLAP uses exactly the same language instructions as our method, generated with the pipeline in Appendix~\ref{appendix:instruc_prompts}, so both methods receive identical language supervision. During LIBERO post-training, CLAP uses the original task instructions; the rule-based action-intent annotations are specific to our intent interface and are not used by CLAP. All methods are evaluated with the same protocol of 50 trials per task.

\begin{table}[h]
\centering
\renewcommand{\arraystretch}{1.3}
\setlength{\tabcolsep}{5pt}
\caption{Comparison with CLAP on the LIBERO benchmark (success rate, \%). All methods share identical pre-training data (HOI4D, TACO, Bridge-V2) and are post-trained separately on each suite. $^\dagger$Re-implemented by us under this protocol. UniVLA results are taken from Table~\ref{table:simulation_libero}, which also lists the remaining baselines.}
\label{table:appendix_clap}
\begin{tabular}{l|cccc|c}
\hline
Methods & Spatial & Object & Goal & Long & Avg. \\
\hline
UniVLA$^\dagger$~\cite{univla} & 90.1 & 91.3 & 88.3 & 82.7 & 88.1 \\
CLAP$^\dagger$~\cite{zhang2026clap} & 92.3 & \textbf{95.6} & \textbf{91.9} & 83.7 & 90.9 \\
\hline
Ours & \textbf{93.4} & 94.2 & 91.2 & \textbf{86.4} & \textbf{91.3} \\
\hline
\end{tabular}
\end{table}

\paragraph{Results and analysis.}
As shown in Table~\ref{table:appendix_clap}, our method outperforms CLAP on two of the four suites and achieves a higher average success rate (91.3\% vs.\ 90.9\%), while remaining competitive on LIBERO-Object and LIBERO-Goal. We attribute these gains to the cleaner action representation learned by IGRD. CLAP derives latent actions from visual dynamics and aligns them with a robot action codebook. Although it introduces an environment codebook, this separation is driven only by reconstruction and has no semantic guidance, so scene-specific factors such as background, object appearance, and layout can still leak into the latent actions. In contrast, IGRD uses explicit intents to guide the disentanglement of the implicit action intent, yielding a representation that keeps the action-relevant information shared across humans and robots and filters out scene-specific noise.

% Check whether the conference requires a reproducibility checklist to be included in the paper.
% If so, you can uncomment the following line and ajust the path to include it.
% \input{ReproducibilityChecklist.tex}

\end{document}